\documentclass{IET}
\usepackage{tikz}
\usetikzlibrary{shapes,arrows,backgrounds,fit,positioning,snakes}

\begin{document}

\title{Segmentation of Large Images Based on Super-pixels and Community Detection in Graphs}

\author[1]{Oscar A. C. Linares}
\affil{Instituto de Ci\^{e}ncias Matem\'{a}ticas e de Computa\c{c}\~{a}o, Universidade de S\~{a}o Paulo, Campus de S\~{a}o Carlos, Caixa Postal 668, 13560-970, S\~{a}o Carlos, Brazil.}

\author[2]{Glenda Michele Botelho}
\affil{Universidade Federal do Tocantins, Palmas, Brazil.}

\author[1]{Francisco Aparecido Rodrigues}
\author[1,*]{Jo{\~a}o Batista Neto}

\affil[*]{jbatista@icmc.usp.br}

\abstract{
Image segmentation has many applications which range from machine learning to medical diagnosis. In this paper, we propose a framework for the segmentation of images based on super-pixels and algorithms for community identification in graphs. The super-pixel pre-segmentation step reduces the number of nodes in the graph, rendering the method the ability to process large images. Moreover, community detection algorithms provide more accurate segmentation than traditional approaches, such as those based on spectral graph partition. We also compare our method with two algorithms: a) the graph-based approach by Felzenszwalb and Huttenlocher and b) the contour-based method by Arbelaez. Results have shown that our method provides more precise segmentation and is faster than both of them.}

%\keywords{Image segmentation \and Community detection \and Super-pixels}

\maketitle

\section{Introduction} \label{sec:Introduction}

Segmentation is one of the most important subjects in image analysis~\cite{Fu1981,Haralick85,Pal1993,Zhang08}. The goal of segmentation is to locate objects and boundaries (e.g. lines and curves) in images according to colour or texture. This task has several practical applications, such as medical diagnosis, machine vision, content-based image retrieval, industrial inspection, materials science experiments, astronomical observations and security screening~\cite{Costa00,Parker010}. Since terabytes of images are generated from such applications, the extraction of quantitative information only makes sense if performed automatically, which makes image segmentation  indispensable.

Algorithms for image segmentation can be grouped according to some underlying approaches, such as image threshold, hybrid linkage region growth, spatial clustering, Markov random fields and neural networks~\cite{Haralick85,Pal1993}. Many surveys on image segmentation methods have been published (e.g.~\cite{Fu1981,Pal1993}). All of these methods are strictly related to the concept of cluster identification in machine learning~\cite{Theodoridis}, whose purpose is to assign a label to every object (or pixel, in the case of images) such that similar objects have the same label. Thus, methods developed for data clustering have been largely applied to image segmentation~\cite{Jain99}. 

Recently, data clustering has been ad\-dres\-sed by concepts and methods of complex networks theory~\cite{Granell011:Chaos,Arruda011, Zanin016}. Complex networks are non-trivial graph, presenting very irregular structure~\cite{Newman10} and modular organization~\cite{Fortunato2016}. Complex networks are made up of elements connected according to a defined type of interaction. For instance, the Internet is formed by routers linked by cables and optical fibres; our brain is formed by a set of neurons connected by synapses; society consists of people linked by social relationships; and protein networks are made up of proteins connected by physical interactions~\cite{Costa011:AP}. Most real-world networks are structured into communities, i.e. sets of nodes densely connected to each other, but sparsely connected with nodes in o\-ther gro\-u\-ps~\cite{Newman10}. The identification of communities can be viewed as a data clustering problem, since methods for network partition require no prior knowledge about number of communities and number of nodes in each community~\cite{Fortunato010,Newman10, Fortunato2016, Zanin016}. Methods implemented for community identification can be taken into account in data clustering when the dataset is structured as graph~\cite{Arruda011, Zanin016}. In this case, a data set is mapped as a network and communities obtained, each representing a cluster~\cite{Granell011:Chaos,Arruda011}. Community identification algorithms have provided very accurate results in data clustering, overcoming traditional methods (e.g.~\cite{Fortunato010,Granell011:Chaos,Arruda011, Fortunato2016}).

Looking at the identification of objects in an image as a data clustering issue, community identification algorithms can also be used to segment images. However, this approach has a practical limitation. Most community identification algorithms scale with the network size~\cite{Fortunato010}. For instance, one of the fastest approaches based on greedy optimization runs in $O(N \log^2 N)$, where $N$ is the number of nodes~\cite{Clauset04:PRE}. An image of size $N\times N$ is mapped as a graph with $N^2$ nodes, since each node represents a pixel. Therefore, in practice, community identification algorithms can deal only with small images.

To overcome this limitation we consider the super-pixel approach to compact images and reduce the number of nodes in the respective network. Each node in the network represents a group of pixels (i.e. a super-pixel), as opposed to a single pixel. That reduces the network size and enables the segmentation of large images. Thus, methods for community identification can be applied and the object identified as communities in the network. We perform a statistical analysis over 120,000 segmented images to define the best parameters in the super-pixel and community detection methods. An adaptive approach for computing the threshold value employed in the community detection is also introduced here. We verify experimentally that our method is more accurate and faster than two other approaches: i) the also graph-based algorithm by Felzenszwalb and Huttenlocher~\cite{Felzenswalb} and ii) Arbelaez's contour-based method \cite{arbelaez2011contour}.

This paper is organized as follows: Section~\ref{sec:RelatedWork} reviews previous investigations related to our image segmentation approach. Concepts of super-pixels, community detection in graphs and segmentation evaluation are discussed in Section~\ref{sec:Concepts}. The methodology is presented in Section~\ref{sec:Methodology}. Results and discussion are in Section~\ref{sec:Experiments}, where a thorough analysis of parameter values is provided  and comparison with the above-mentioned image segmentation methods is presented. Finally,  conclusions are addressed in Section~\ref{sec:Conclusion}. 	

\section{Related work} \label{sec:RelatedWork}

To perform graph--based image segmentation, pixels are represented as nodes and pairs of pixels are connected according to the similarity between their features. An image of size $N\times N$ can be mapped onto a graph with $N^2$ nodes. For large values of $N$, which is the case of most real size images~\cite{Howe08}, both computational cost and memory constraints can be a crucial drawback.

Many graph-based methods can be used for image segmentation~\cite{Wu,Cox,Shi,Wang,Felzenswalb}. Some of these approaches find a set of edges whose removal disconnects the graph, creating isolated groups of vertices (called components)~\cite{Wu,Cox,Shi,Wang}. These methods are based on the optimization of a cost function. A cost function is the total weight of the removed edges with an additional value that penalizes the creation of small components. Wu and Leahy~\cite{Wu} considered a minimum cut criterion, which seeks to partition a graph into $k$ subgraphs by minimizing the sum of the edge weights. The minimum criterion favours the creation of small components by additional parameters~\cite{Cox,Shi,Wang}. Cox et al.~\cite{Cox} considered the dimension of the components to balance the partitioning (ratio cut). Wang and Siskind~\cite{Wang} used the number of cut edges (minimum mean cut). Shi and Malik~\cite{Shi} presented a method, called normalized cut, which considers the sum of the node degrees of each partition to balance the partitioning. The high computational cost of these approaches makes them unattractive to applications that involve large images. In addition, such graph partition algorithms do not provide the most accurate results, since recent methods for community identification revealed to be more accurate than classical methods, such as those based on spectral bisection~\cite{Fortunato010,Newman10}.

Felzenszwalb and Huttenlocher~\cite{Felzenswalb} suggested an approach for image segmentation derived from a pairwise region comparison. The method defines a criterion to evaluate whether there exist an edge between two regions based on a greedy strategy and uses a graph representation of the image to obtain the final segmentation. Three parameters are required: i) the minimum component size, ii) the Gaussian smoothing factor and iii) a threshold value that controls the similarity between pixels. The method is almost linear-time in the number of graph edges and can be employed for the segmentation of large images. Since it falls into the same category as our proposed approach, we also compare our framework with this algorithm in Section~\ref{sec:Experiments}.

\section{Concepts} \label{sec:Concepts}

The techniques presented in this paper combines algorithms for communities identification in graphs and the super-pixels approach to segment large images. A brief overview of the concepts that underpin our method is provided as follows.

\subsection{Super-pixels} \label{sec:SuperPixel}

Super-pixel is a region-based image segmentation approach to over-segment the image by grouping pixels that belong to the same object~\cite{Ren,Levinshtein,Cigla,Achanta,Achanta2}. In this paper, we will consider super-pixels extraction techniques that start with a regular grid of arbitrary size placed over an image. By means of an iterative process, the edges of such grid eventually converge to the actual boundaries of objects (see Fig.~\ref{Fig:Superpixels}). Recent papers \cite{Cigla,Achanta2} have proposed techniques based on $k$-means clustering algorithm, that provide accurate results and low computational cost --- $O(N)$, where $N$ is the number of pixels. These techniques are presented below.

\subsubsection{Speeded-up turbo pixels} \label{sec:SUTP}

\begin{figure}[!htb]
\centering
  \subfigure[]{\fbox{\includegraphics[width=.30\linewidth]{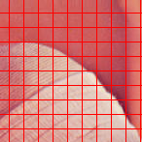}}}
  \subfigure[]{\fbox{\includegraphics[width=.30\linewidth]{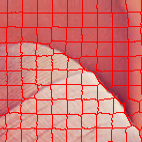}}}
  \subfigure[]{\fbox{\includegraphics[width=.30\linewidth]{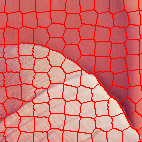}}} \\
  \caption{Super-pixels computation: (a) initial regular grid; (b) grid after three iterations and (c) final grid after ten iterations.}
 \label{Fig:Superpixels}
\end{figure}

Cigla and Alatan~\cite{Cigla} suggested an efficient method for super-pixel extraction, called speeded-up turbo pixels (SUTP), which yields quasi-uniform super-pixels. The method is efficient in terms of computational costs, compactness of segments and over-segmentation errors~\cite{Cigla}. The following pseudo-code describes the SUTP algorithm:
\begin{enumerate}
\item Initially, a regular grid divides the image into rectangular regions, or segments (Fig. \ref{Fig:Superpixels}(a)).
  \item Pixels at the over-segmented boundaries are tested and assigned to new segments by minimizing a cost function, defined as:
    \begin{equation}
        \label{eqn:Superpixel}
	C_{x,y}(i)=\lambda_{1}\vert I(x,y)-I_{i}\vert +\lambda_{2}\vert (x-C_{x}^{i})^{2}+(y-C_{y}^{i})^{2}\vert
 \end{equation}
    \noindent where $I_i$ is the mean intensity of the $i$-th segment, $x$ and $y$ are the coordinates of the pixel tested among different segments, $C_x^i$ and $C_y^i$ are the centroids of the $i$-th segment and $\lambda_1$ and $\lambda_2$ correspond to the intensity weigth similarity and convexity constraints, respectively. The first term ensures that pixels of similar colours are merged, whereas the second enables super-pixels to have more uniform and convex shapes (rigidity). The higher the $\lambda_2$, the stiffer the super-pixel border will be. The lower, the more flexible. A similar behaviour applies to $\lambda_1$, regarding the importance given to colour during segmentation. Once all the boundary pixels have been tested, super-pixel mean intensity and centre positions are updated, yielding new super-pixels.  
    
 \item This iterative process stops when the number of interchanged pixels has reached a given threshold. At this stage, super-pixels are said to have converged, i.e. the initial regular grid cells become irregular and their borders match the object boundaries in the image, as shown in Fig.~\ref{Fig:Superpixels}(c).
\end{enumerate}

Parameters $\lambda_1$ and $\lambda_2$ have been extensively analysed here. As colour should be highly weighted during segmentation,  $\lambda_1$ is always set to 1. We consider $\lambda_2 \ll \lambda_1$ to get super pixels well adjusted to the actual contours found in images. 

\subsubsection{Simple linear iterative clustering} \label{sec:SLIC}

Achanta et al.~\cite{Achanta,Achanta2} introduced an iterative algorithm for super-pixels extraction of colour images based on CIELAB colour model. Like SUTP, this simple linear iterative clustering (SLIC) approach is  based on the $k$-means algorithm, but it employs a more efficient search space which implies a reduced computational cost. The  SLIC approach is described as follows:

\begin{enumerate}
  \item An input image with $N$ pixels is partitioned into $k$ rectangular regions whose dimensions may vary slightly, i.e. super-pixels of expected dimensions $S \times S$, where $S = \sqrt{\frac{N}{k}}$. A cluster center, i.e. a 5-dimensional vector  $C_j = \{L_j, a_j, b_j, x_j, y_j\}$ represents each super-pixel $j$. $L_j$, $a_j$ and $b_j$ are the mean values of the three CIELAB components of super-pixel $j$. The lightness $L$ ranges from 0 (black) to 100 (white) and $a$ and $b$ are the green--red and blue--yellow colour components, respectively. The centroid of super-pixel $j$ is given by $(x_j,y_j)$.

  \item At each iteration, each pixel $i$ is associated with the super-pixel with the closest centres, given that the search area of such super-pixel includes $i$. The size of this search area is $2S\times 2S$. Then, new cluster centres are computed by taken the mean value of vector $C_j$, considering all pixels in super-pixel $j$. This process corresponds to the minimization of function $D$, defined as
  \begin{equation} \label{eqn:Slic}
     D = \sqrt{d_c^2 + \left(\frac{d_s}{S}\right)^2 m^2},
\end{equation}
where $d_c$ is the distance in the colour space,
\begin{equation}
d_c = \sqrt{(l_j - l_i)^2 + (a_j - a_i)^2 + (b_j - b_i)^2},
\end{equation}
and $d_s$ is the spatial distance, i.e.
\begin{equation}
        d_s = \sqrt{(x_j - x_i)^2 + (y_j - y_i)^2}.
\end{equation}
Achanta et al. suggested that accurate super-pixels can be obtained with at most 10 iterations~\cite{Achanta}.

  \noindent Parameter $m$ ($1\leq m \leq 20$) of Eq.~\ref{eqn:Slic} controls the super-pixels compactness. The higher the value of $m$, the greater the emphasis on spatial proximity and the more compact the cluster~\cite{Achanta}.

  \item Finally, isolated pixel groups can arise after the iterative process. These groups should be associated with the largest super-pixel in their neighbourhood.
\end{enumerate}

\subsection{Images as complex networks} \label{sec:ImagesForGraphs}

Images can be represented as graphs (or complex networks), in which pairs of pixels are connected according a similarity function. The most common functions to quantify the similarity between pixels are based on Euclidean, Manhattan or Gaussian distances (e.g.~\cite{Felzenswalb}). Such functions generally take into account two parameters, namely threshold ($t$) and radius ($R$). Parameter $t$ determines the number of connections in the graph. Pixels $i$ and $j$ are connected if the strength of their similarity (connection weight) is lower than or equal to $t$. Additionally, radius $R$ delimits a circular region which encompasses all pixels that can be connected to each other.

In this paper, nodes in the graphs are super-pixels, instead of pixels. Many colour--, shape-- or texture--based features or descriptors can be computed from super-pixels. We have tried histograms, Local Binary Patterns (LBP) and statistical moments. However, best performance was achieved with a  3-dimensional descriptor formed by the components of the CIELAB colour space (the mean CIELAB value of all pixels belonging to a super-pixel). This simple descriptor was not only faster to compute, but also more discriminant than texture--based counterparts. We employ the Euclidean distance to compute the similarity between two super-pixels. Super-pixels will be connected if similarity  is bellow or equal a certain threshold, which can be defined by the adaptive approach proposed here (Section~\ref{subs:GraphGeneration}). 

\subsection{Community identification in graphs} \label{sec:CommunityIdentification}

Most complex networks present community structure (modular organization)~\cite{Costa011:AP, Fortunato2016}. Communities can be defined as groups of nodes that are more densely linked to each other than to the rest of the network~\cite{Girvan02}. The identification of communities in networks can be understood as the clustering task in data mining~\cite{Arruda011, Zanin016}.

The number of communities and their sizes are generally unknown~\cite{Newman10}. To quantify the quality of a particular division of the network, Newman proposed a measure called modularity. For a network partitioned into $m$ communities, a matrix $E$, $c \times c$ is defined and its elements $e_{ij}$ represent the fraction of connections between communities $i$ and $j$. Modularity $Q$ is calculated as
\begin{equation}\label{Eq:modularity}
Q = \sum_i [ e_{ii} - ( \sum_j e_{ij} )^2 ] = \mathrm{Tr} E  - ||E^2||,
\end{equation}
where $0 \leq Q < 1$.  In practice, networks with well defined modular structure exhibit $0.3 \leq Q \leq 0.7$~\cite{Newman10}.

\subsubsection{Fast greedy algorithm}

Fast greedy is a hierarchical agglomerative algorithm for community detection based on the maximization of the modularity measure~\cite{Clauset04:PRE}. It improves on the algorithm proposed by Newman~\cite{Newman04} which uses a greedy optimization procedure. The algorithm starts with each vertex representing an isolated community. The method repeatedly joins communities into pairs by choosing, at each step, the merging that results in the largest increase (or smallest decrease) of $Q$. The best division corresponds to the partition that resulted in the highest value of $Q$.

The computational cost of the algorithm proposed by Newman~\cite{Newman04} is $O((M + N)N)$ or $O(N^2)$ for sparse graphs, where $N$ is the number of nodes and $M$ is the number of edges. Due to efficient data structures, the computational cost drops to $O(M d (\log N))$~\cite{Clauset04:PRE}, where $d$ corresponds to the depth of the dendrogram that describes the hierarchical structure of the network. For sparse and hierarchical networks, $M\sim N$ and $d\sim \log N$. Therefore, the fast greedy runs in $O(N\log^2N)$.

\subsection{Evaluation of image segmentation} \label{sec:SegmentationEvaluation}

The metric proposed by Arbelaez et al.~\cite{Arbelaez} is usually employed to quantify the similarity between two segmented images, which is based on the number of pixels in the intersections of the regions that form each image. The overlap between regions $R$ and $ R'$  is defined as
\begin{equation}
    O(R,R') = \frac{|R  \cap R'|}{|R \cup R'|}.
\end{equation}
The covering of a segmentation $S$ by a segmentation $S'$ is calculated by
\begin{equation}
    C(S,S') = \frac{1}{N} \sum_{R\in S} |R| \max_{R' \in S'} \{O(R,R')\},
\end{equation}
\noindent where $N$ denotes the total number of pixels in the image. Function $\max(\cdot)$ returns the maximum value of the overlap between regions $R$ and $R'$. Note that this formulation does not satisfy the axiom of symmetry, $C(S,S')\neq C(S',S)$.

To overcome this limitation, we use a metric called \textit{Adjustable Object-oriented Measure (AOM)}, proposed by Cuadros Linares et al. \cite{cuadros_measure} to quantify the similarity between two segmented images. Consider two segmented images $S$ and $S'$ composed of $n$ and $n'$ regions, respectively. Matrix $M$ of size $n \times n'$ is defined as
\begin{equation}     \label{Eqn:metricaMatriz}
    M_{ij} = |R_i \cap R'_j|,
\end{equation}
\noindent where $R\in S$ and $R' \in S'$.

Intersection $I$ between two segmented images $S$ and $S'$ is calculated by

%\begin{equation} \label{eqn:IntersectionMetric}
%    I(S,S')= \frac{1}{N}\sum_{k=1}^{min(n,n')} \max_{k}(M),
%\end{equation}

\begin{equation} 
\label{eqn:IntersectionMetric}
I(S,S')= \frac{1}{N}\sum_{k=1}^{min(n,n')} \max_{k}(M)\mathrm{penalty}(k),
\end{equation}
\noindent where $N$ is the number of pixels in the image and $\max_{k}(M)$ returns the $k$-th largest element of $M$ and the function $\mathrm{penalty}(\cdot)$ gives the corresponding over-segmentation penalty values for all $k$ largest intersections in $M$. Over-segmentation occurs when a region $R_i$ has a number $s$ ($s\geq2$) of corresponding regions $R'$ ($R_i \cap R'_j > 0$). Then, the penalty is computed as: 

\begin{equation}
\mathrm{penalty}(k) = 
\begin{cases}
\frac{1}{\alpha s},& \text{if } \alpha s \geq 1\\
1,              & \text{otherwise}
\end{cases},
\end{equation}

\noindent in which $\alpha \in [0,1]$ is a parameter to indicate the over-segmentation penalty and $s$ is the number of regions $R'$ corresponding to the region $R_i$, where $R_i \cap R'_j > 0$. AOM satisfies the following properties:
\begin{enumerate}
    \item $ 0 \leqslant  I(S, S')  \leqslant 1.$
    \item $I(S, S') = 1$, if $S=S'.$
    \item $I(S, S') = I(S', S).$
    \item $I(S, S') \geq I(S, S'')  \And  I(S, S'') \geq I(S', S'')  \Longrightarrow I(S, S') \geq I(S', S'').$
\end{enumerate}

\section{Methodology} \label{sec:Methodology}

\begin{figure}[!t]
\centering
\begin{tikzpicture}[]
\centering
\tikzstyle{block}    = [rectangle, draw=white, thick, text width=7em, text centered, rounded corners, minimum height=3.75em]
\tikzstyle{line}     = [draw, -latex',  thick]
\matrix [column sep=8mm,row sep=9mm]
  {
      % row 1

      \node [block](superpixel_image)
      {
	\centering
	\includegraphics[width=6.3em, height=4.88em]{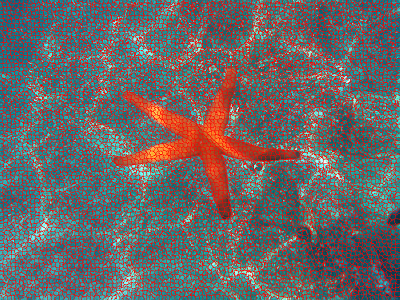}
	Super-pixels
      };
      &
      &
      &
      \node [block](complex_network)
      {
	\centering
	\includegraphics[width=6.3em, height=4.88em]{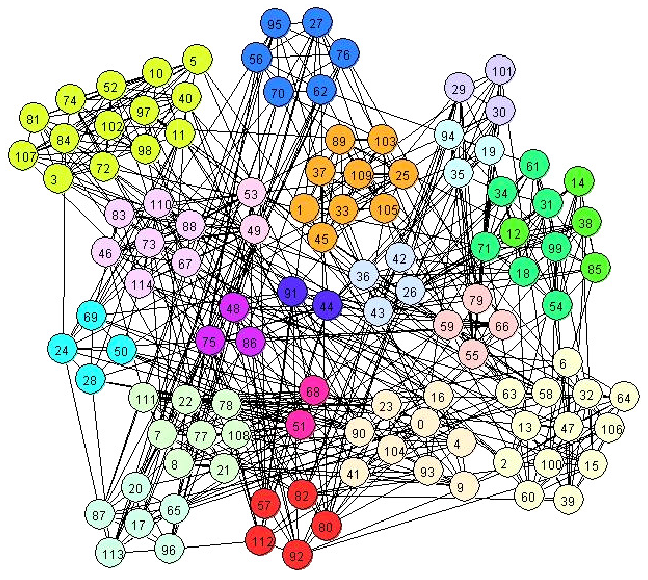}
	Graph
       };
      \\
      %row2
      \node [block](original_image)
      {
	\centering
	\includegraphics[width=6.3em, height=4.88em]{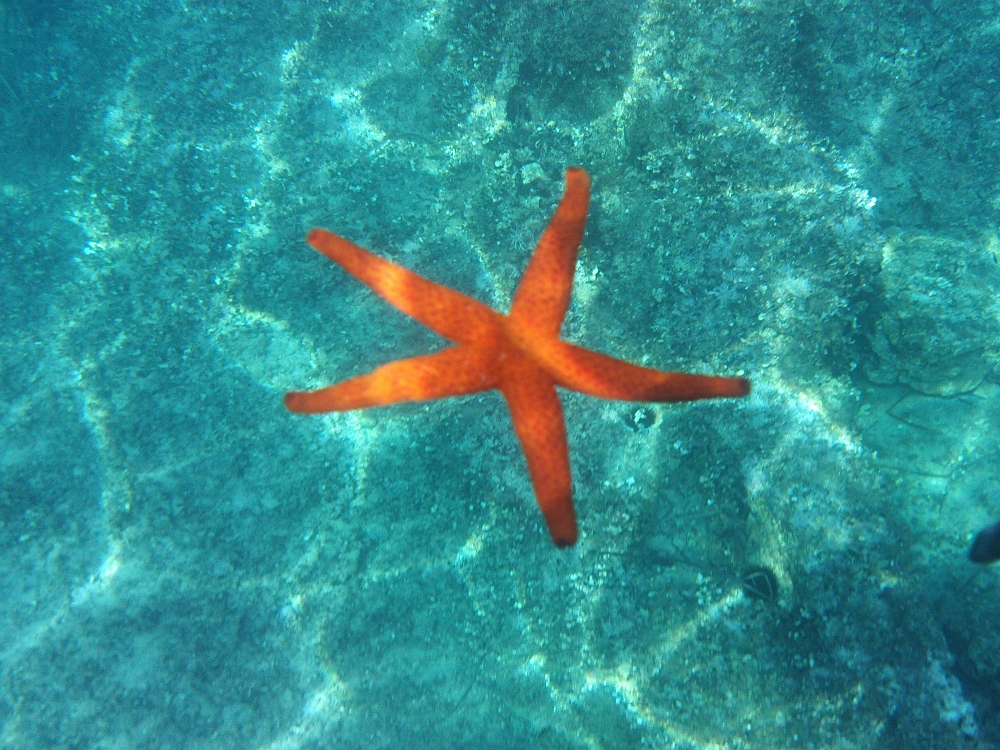}
	Original image
      };
      &
      &
      &
      \node [block](segmented_image_0)
      {
	\centering
	\includegraphics[width=6.3em, height=4.88em]{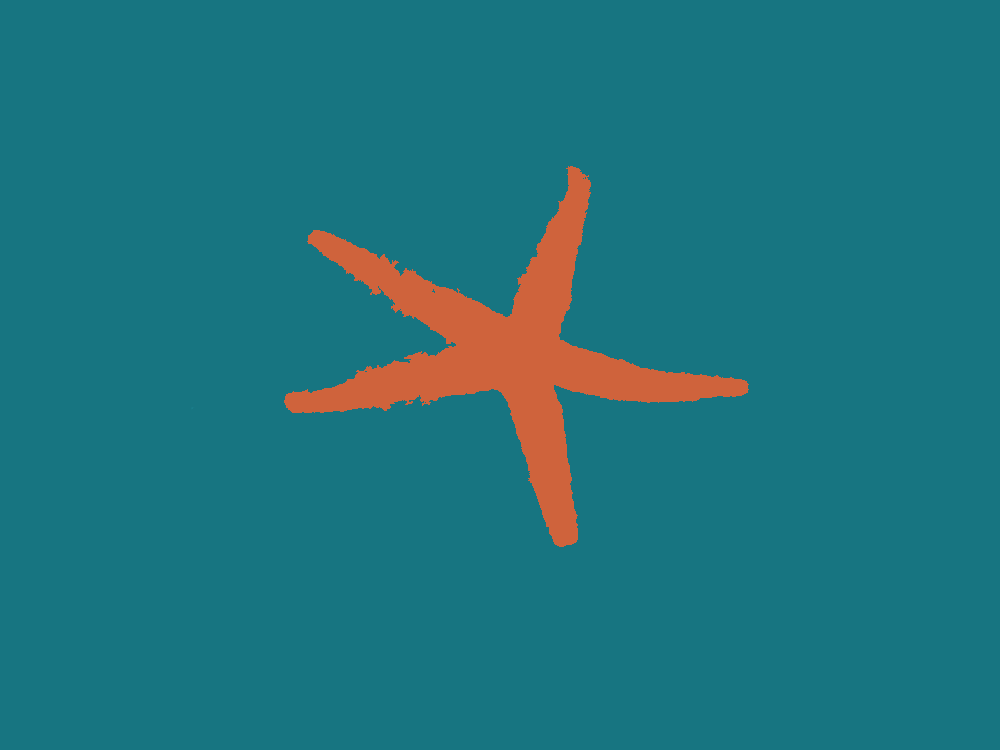}
	Segmented image
      };\\
};

  \begin{scope}
    \path [line] (original_image)   edge node[left] {\footnotesize{}} (superpixel_image);
    \path [line] (superpixel_image) edge node[auto] {\footnotesize{Graph generation}}       (complex_network);
    \path [line] (complex_network) edge node[left] {\footnotesize{Community detection}} (segmented_image_0);
  \end{scope}

\end{tikzpicture}
\caption{Image segmentation approach based on community identification combined with super-pixels. Super-pixels are extracted from the original image and a graph is created. Community detection algorithms provide the segmented image.} 	
\label{Fig:Methodology}
\end{figure}
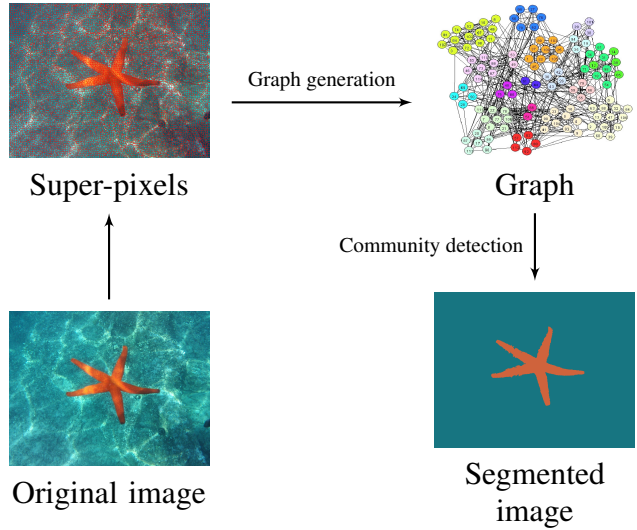

Fig.~\ref{Fig:Methodology} depicts the proposed image segmentation framework. Initially, a simple or quadtree grid covers the image. Next, the super-pixel algorithm by Cigla and Alatan~\cite{Cigla} produces super-pixels with irregular cells, whose borders match the boundary of objects in the image. We model the image as a graph and assign each community to an object. Finally, we quantify the accuracy of the segmentation with the method described in Section~ \ref{sec:SegmentationEvaluation}. Each of these tasks is detailed in the following sub-sections.

\subsection{Super-pixel extraction}

We chose the algorithm by Cigla and Alatan~\cite{Cigla} instead of SLIC~\cite{Achanta,Achanta2},  since the latter is slower than the former. SLIC analyses the entire region of a super-pixel to move the boundary position, whereas the algorithm by Cigla and Alatan takes into account only the pixels along the boundaries~\cite{Cigla}. Moreover, the algorithm by Cigla and Alatan enabled the transition from the original regular grid to a quadtree-based approach, first proposed in this paper. As shown in Section~\ref{sec:Experiments}, the quadtree-based grid yields better results with shorter processing time than the regular grid for images with large and uniform regions.

We modified the original algorithm so as to consider the colour model CIELAB in the convergence Function \ref{eqn:Superpixel} as opposed to pixel intensity. Hence, this convergence function takes into account the three CIELAB components: $L$ (lightness, which varies from 0 (black) to 100 (white)); $a$, representing colours from green to red and $b$, colours from blue to yellow.

Four distinct parameters must be dealt with: (i) the size of the initial squared grid ($s$), (ii) the number of iterations ($i$), (iii) the colour weight $\lambda_1$ and (iv) the rigidity weight $\lambda_2$. Changes in parameters $s$ and $i$ can affect both speed and quality of the super-pixels produced (border matching). The bigger the initial super-pixel size $s$, the larger the number of iterations required for convergence. However, as we experimentally described in a previous work~\cite{Cuadros012}, only 10 iterations are necessary for a correct super-pixels convergence --- in fact, only six iterations are enough for most cases. We have also shown that parameter $s$, defined as $10\times 10$, produced the best results for both speed and correct convergence. Following the discussion presented in Section~\ref{sec:SUTP}, we set   $\lambda_1 = 1$ and $\lambda_2 = 0.1$ in our experiments.

%The influence of parameter $\lambda_1$ on the segmentation quality is significantly reduced when the Euclidean distance is adopted and the colour model is changed from RGB to CIELAB. Moreover, the best segmentation results for the Berkeley dataset are obtained with $\lambda_1=1$. Regarding $\lambda_2$, higher values generate more convex super-pixels, which are difficult to adjust to the edges of image objects. Therefore, lower values for parameter $\lambda_2$ are more suitable for image segmentation. Our experiments show that values of $\lambda_2$ ranging from 0.1 to 0.5 produce accurate results.

\subsection{Graph generation} \label{subs:GraphGeneration}

After the super-pixel extraction, we build a graph in which nodes represent super-pixels connected according to the following weight function

\begin{equation}
   \label{eqn:CIELABIntensity}
    W_{i,j} = \sqrt{(L_i - L_j)^2 + (a_i - a_j)^2 +(b_i - b_j)^2},
\end{equation}

\noindent where $i$ and $j$ are two graph nodes (super-pixels) and  $L$, $a$ and $b$ are the CIELAB channels as previously explained. The nonlinear relations between $[L,a,b]$ aim to simulate the nonlinear behaviour of the human eye. Also, uniform changes in perceived colours reflect in uniform changes of the  $L$, $a$ and $b$ components of CIELAB colour space. As a result, perceptual differences between two colours can be approximated by considering each colour as a point in a three-dimensional space, which can be conveniently measured with the Euclidean distance \cite{jain1989fundamentals}. For that reason, we compute the similarity between two super-pixels with the Euclidean distance (Equation \ref{eqn:CIELABIntensity}) by taking the mean $[L,a,b]$ of all pixels belonging to a super-pixel. This is more accurate than using RGB colour models.    
 
A connection is considered only if $W_{i,j} \le t$, where $t$ is a threshold parameter. To avoid an arbitrary choice of $t$, we propose an adaptive thresholding approach. The threshold takes an initial value which gradually increases while the node's degree under evaluation is $\kappa_i \le 0$. After connecting the node $i$, and before evaluating node $i+1$, the threshold is again set to its original value. This strategy prevents community detection algorithms from generating isolated nodes ($\kappa=0$), or a community with a single element. In image processing this is seen as an over-segmented image. Both threshold and increment are initially assigned value $0.5$. This process ends when no isolated nodes are found in the graph or when threshold $t=40$. 	

\begin{figure}[!t]
        \centering
        \includegraphics[scale=0.8]{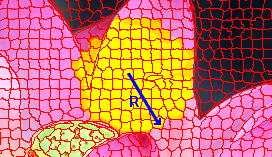}
        \caption{Super-pixel neighborhood (yellow region) of radius $R=5$.}
        \label{Fig:super_pixel_neighborhood}
\end{figure}

Connections between super-pixels are also established only if they are within the same circular region of radius $R$. Fig.~\ref{Fig:super_pixel_neighborhood} shows a yellow region that encompasses all super-pixels within a distance $R=5$ from a given super-pixel. The choice of the value of $R$ is important to avoid connections between two super-pixels that are far apart. 	

\subsection{Image segmentation}

Following the graph construction, community detection algorithms are used to segment images. We have considered the fast greedy method~\cite{Clauset04:PRE}, since it yields accurate results with lower computational cost in comparison with other methods~\cite{Fortunato010}. Fast greedy has been implemented in the igraph library~\cite{Csardi06}. Based on optimization, fast greedy does not ensure the best network division (the algorithm to find the best network division is $NP$-complete), but the outcome is much better than that provided by local methods, such as label propagation~\cite{Cuadros012}. 	

\subsection{Quantitative segmentation evaluation} \label{subs:MetEvaluation}

\begin{figure}[!tb]
\centering
  \subfigure[Original]{\includegraphics[width=.30\linewidth]{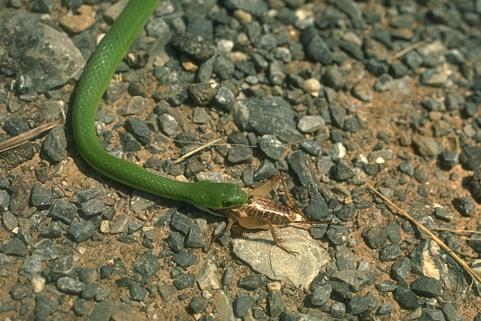}}
  \subfigure[Reference]{\includegraphics[width=.30\linewidth]{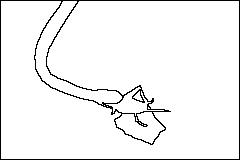}}\\
  \subfigure[]{\includegraphics[width=.20\linewidth]{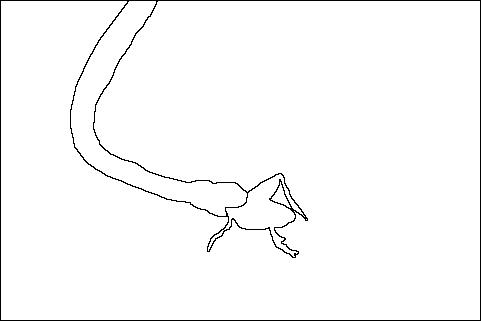}}
  \subfigure[]{\includegraphics[width=.20\linewidth]{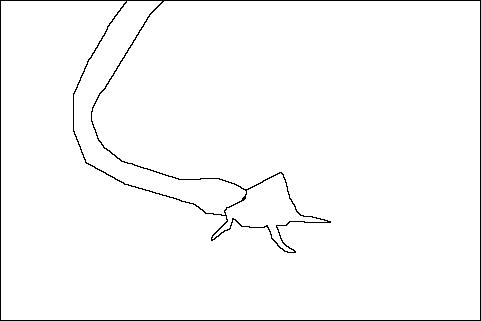}}
  \subfigure[]{\includegraphics[width=.20\linewidth]{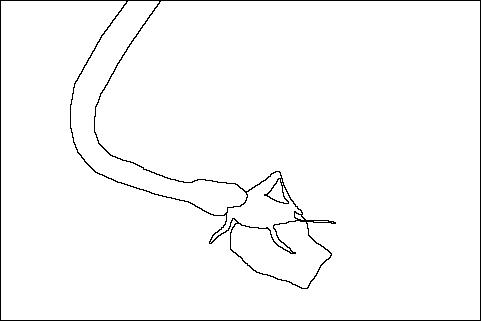}}
  \subfigure[]{\includegraphics[width=.20\linewidth]{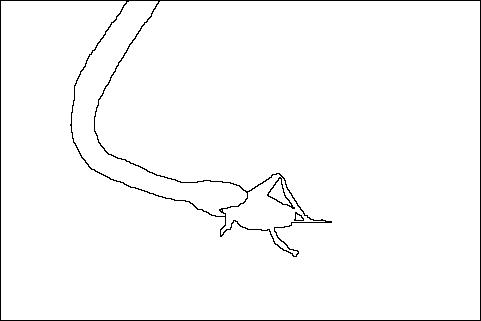}}\\
  \subfigure[Reference]{\includegraphics[width=.20\linewidth]{Cobra6-2.jpg}}
  \subfigure[]{\includegraphics[width=.20\linewidth]{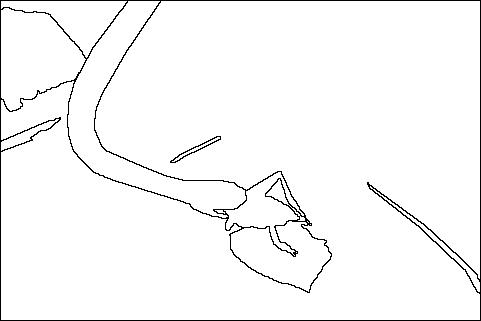}}
  \subfigure[]{\includegraphics[width=.20\linewidth]{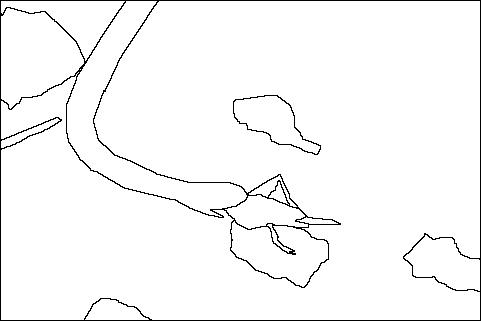}}
  \subfigure[]{\includegraphics[width=.20\linewidth]{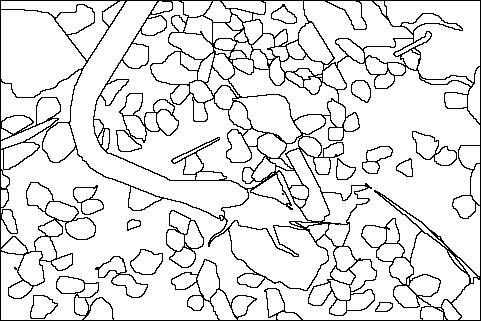}}
  \caption{Proposed reference image selection approach. (a) Original image; (b) selected reference image - same as (g) - according to Eq.~\ref{eqn:IntersectionMetric}; (c)-(j) manual segmentations generated by different users.}
  \label{Fig:ReferenceImage}
\end{figure}

A ground truth image dataset was used for the quantitative evaluation of our automatic segmentation method. The Berkeley University dataset~\cite{ImgBerkeley} consists of 300 natural scene images, each containing a variable number of manual segmentations performed by different subjects. The segmentations of each image vary considerably due to the subjective nature of human perception of image objects. We propose a quantitative approach to choose the reference segmented image among all manual segmentations. For each original image in the dataset, we take all its $n$ manual segmented images and build a square matrix $M$, whose elements quantify the similarity between the manual segmented images $i$ and $j$, i.e. $M_{i,j} = I(i,j)$ (see Eq.~\ref{eqn:IntersectionMetric}). The reference segmented image are the one which yields the maximum value of $\sum_j M_{ij}$ or $\sum_i M_{ij}$. This image can be understood as the most likely human segmentation and is a fair choice for the quantitative evaluation of our methodology, since it is generally desired that segmentations agree with human perception. Fig. \ref{Fig:ReferenceImage} illustrates the reference image selection process.

Although fully automatic, the proposed method must handle six different parameters: four in the super-pixel stage (the size of the initial squared grid ($s$), the number of iterations ($i$), the colour weight $\lambda_1$ and the rigidity weight $\lambda_2$) and two during the graph generation process (threshold $t$ and radius $R$). We have already described in this paper how to set the best values for the four super-pixels parameters so as to make the approach suitable for the segmentation of real large images. In the next section, we show how we defined the best values for parameters $t$ and $R$, used in the graph generation.

\section{Results and discussion} \label{sec:Experiments}

%As discussed in section \ref{subs:GraphGeneration}, our framework for image segmentation based on super-pixels and community identification requires four parameters: $\lambda_1$, $\lambda_2$, radius and threshold. Slight changes in these parameters may considerably affect the outcome of the segmentation~\cite{Cuadros012}. 

The experiments provided in this section address the issues related with the parameters radius and threshold. We ran experiments to show how changes in radius and threshold values affect segmentation. The novel quadtree-based super-pixel is compared with the original implementation, which describes how to attain segmentation at a reduced computational cost. We provide a quantitative evaluation of our technique with Berkeley ground truth dataset. Finally, we compare our approach with Arbelaez's image segmentation method based on contour detection \cite{arbelaez2011contour} and with Felzenszwalb and Huttenlocher's ~\cite{Felzenswalb}, which is also based on graph concepts. 	

\subsection{Using the quadtree approach to generate the initial super-pixel grid}

\begin{figure*}[!htb]
 \centering
  \subfigure[]{\fbox{\includegraphics[width=.3\linewidth]{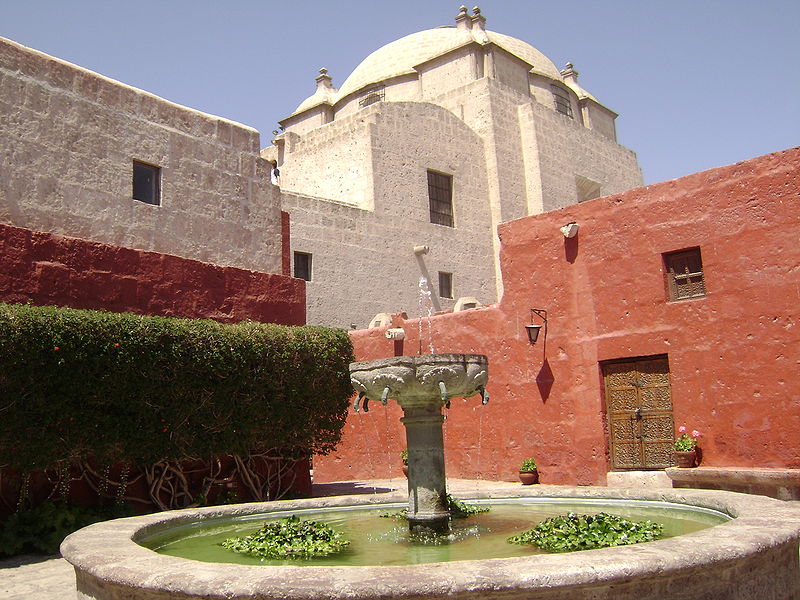}}}\\
  \subfigure[]{\fbox{\includegraphics[width=.3\linewidth]{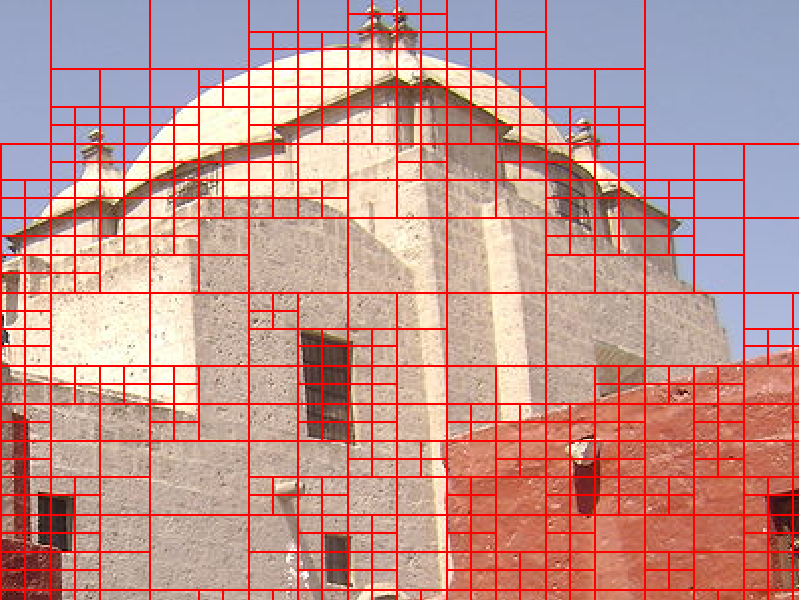}}}
  \subfigure[]{\fbox{\includegraphics[width=.3\linewidth]{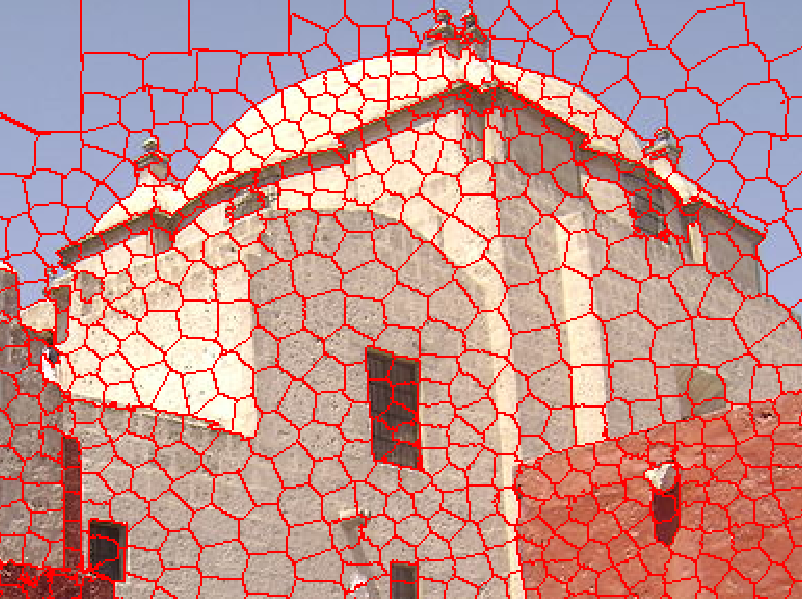}}}
  \subfigure[]{\fbox{\includegraphics[width=.3\linewidth]{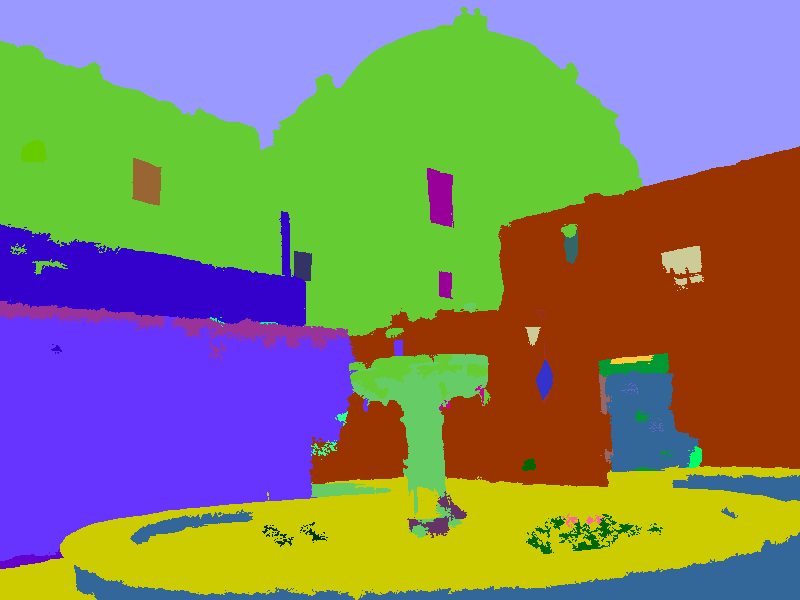}}}\\
  \subfigure[]{\fbox{\includegraphics[width=.3\linewidth]{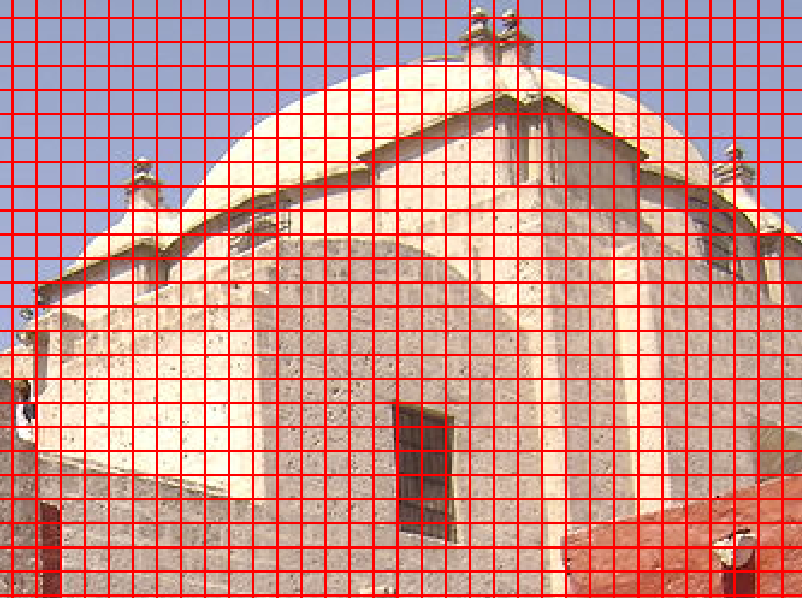}}}
  \subfigure[]{\fbox{\includegraphics[width=.3\linewidth]{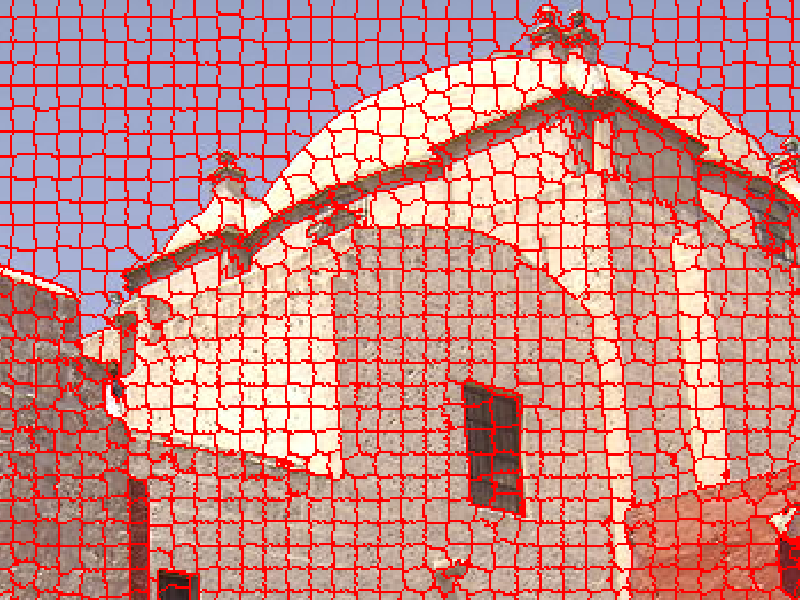}}}
  \subfigure[]{\fbox{\includegraphics[width=.3\linewidth]{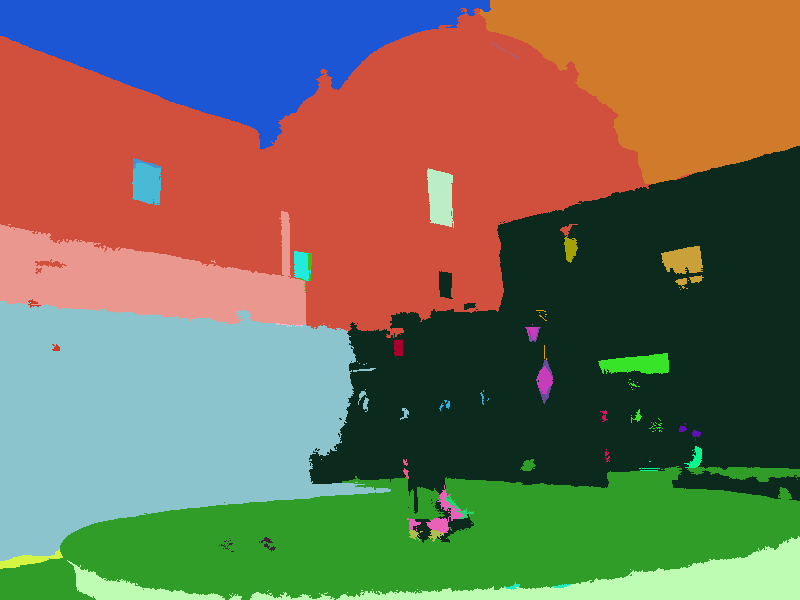}}}
  \caption{Super-pixel segmentation with SUTP and QSTUP. (a) original $800\times 600$ image, (b) quadtree (zoom in), (c) QSTUP (zoom in), (d) segmentation by QSTUP with 30 objects in total, (e) original grid (zoom in), (f) original STUP (zoom in) and (g) segmentation by STUP with 34 objects in total.}
 \label{fig:SantaCatalinaResults}
\end{figure*}

\begin{figure*}[!htb]
 \centering
  \subfigure[] {\fbox{\includegraphics[width=.3\linewidth]{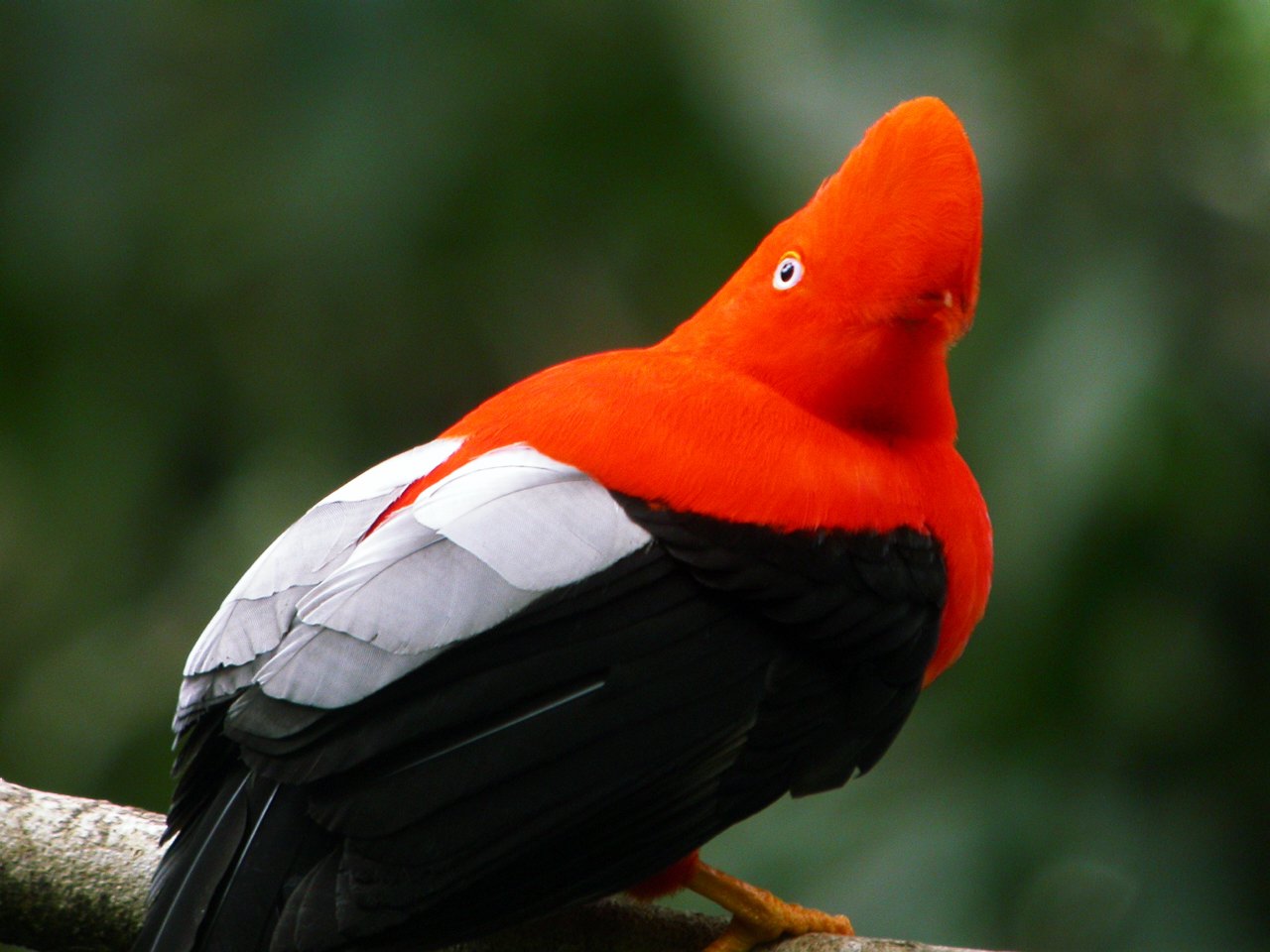}}}\\
  \subfigure[]{\fbox{\includegraphics[width=.3\linewidth]{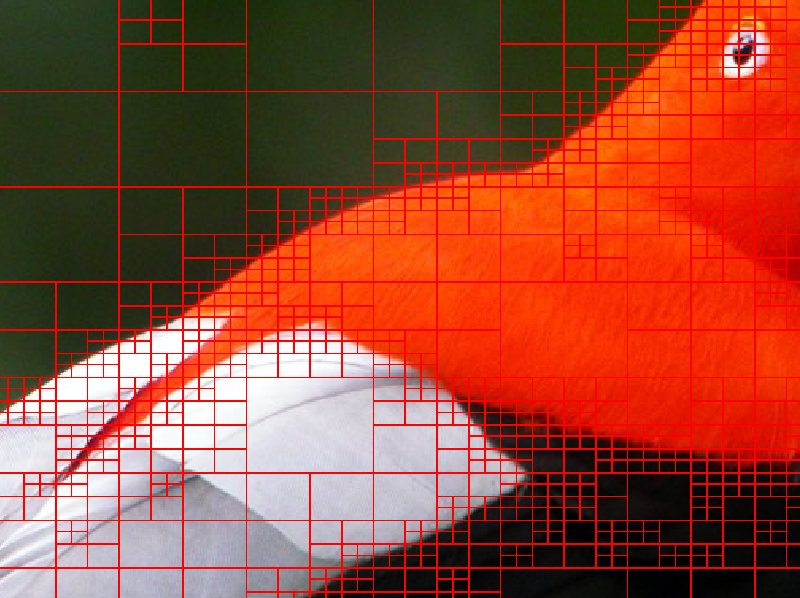}}}
  \subfigure[]{\fbox{\includegraphics[width=.3\linewidth]{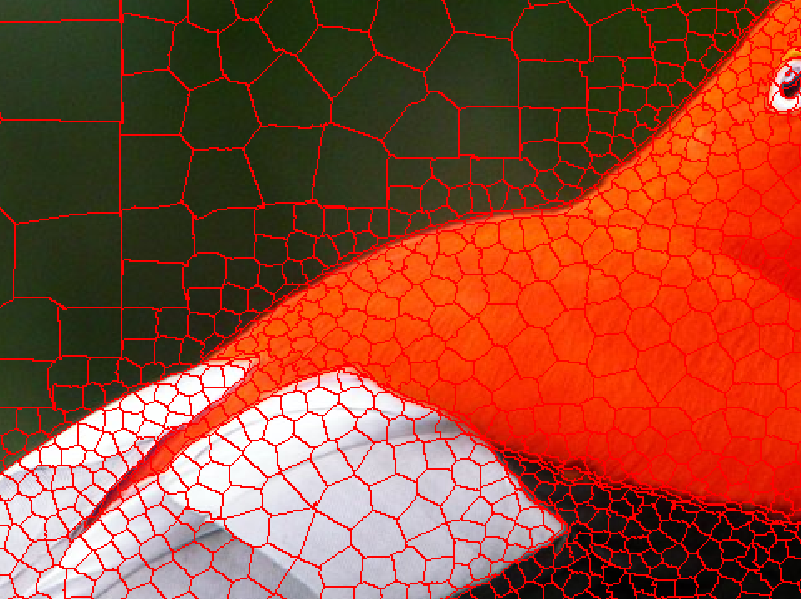}}}
  \subfigure[]{\fbox{\includegraphics[width=.3\linewidth]{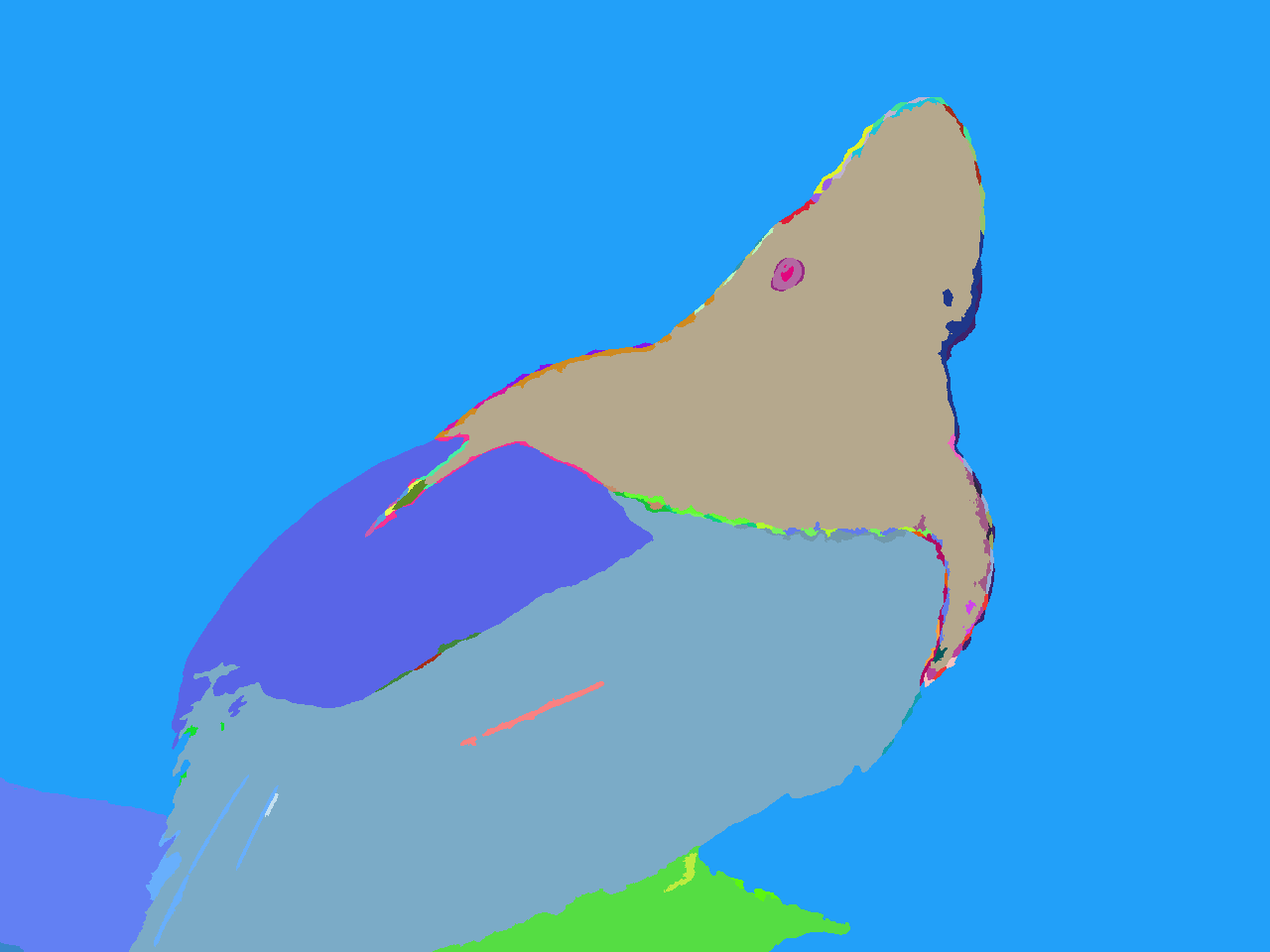}}}\\
  \subfigure[]{\fbox{\includegraphics[width=.3\linewidth]{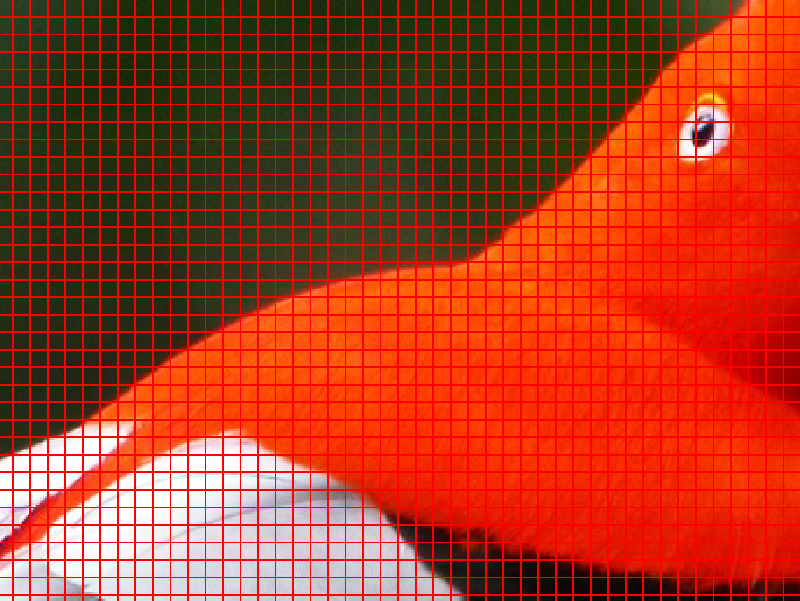}}}
  \subfigure[]{\fbox{\includegraphics[width=.3\linewidth]{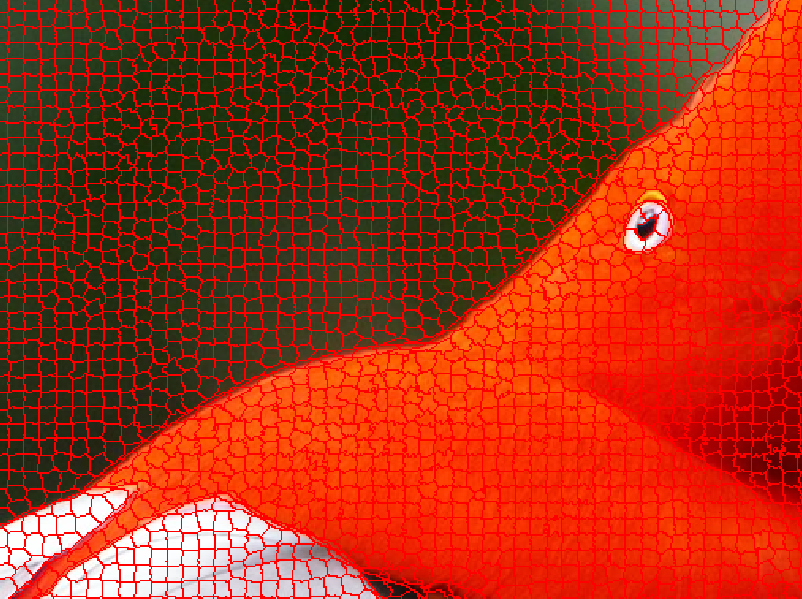}}}
  \subfigure[]{\fbox{\includegraphics[width=.3\linewidth]{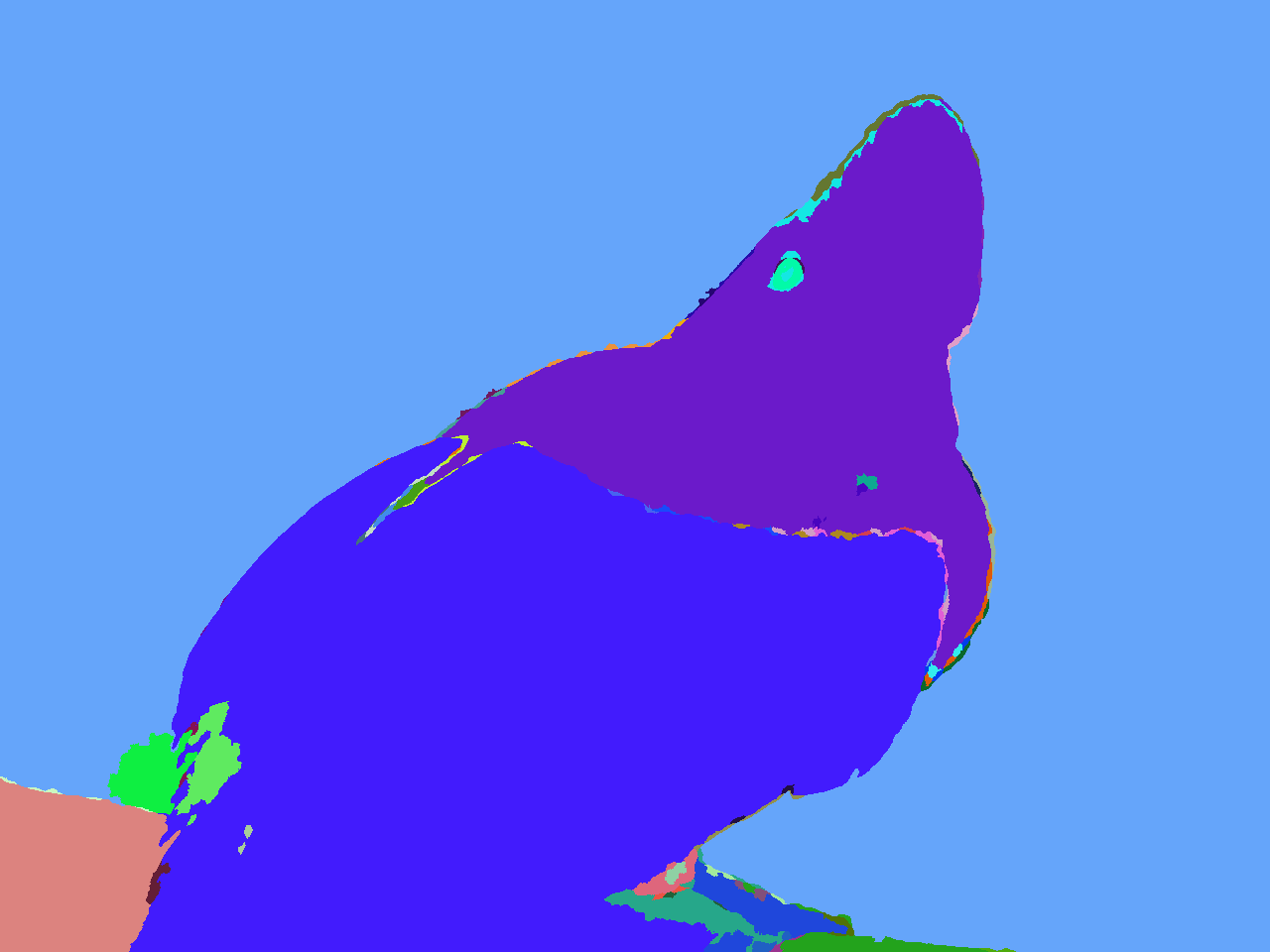}}}
  \caption{Super-pixel extractions  with SUTP and QSTUP. (a) Original $1280\times 960$ image, (b) quadtree (zoom in), (c) QSTUP (zoom in), (d) segmentation by QSTUP with 62 objects, (e) original grid (zoom in), (f) original STUP (zoom in) and (g) segmentation by STUP with 59 objects.}
 \label{fig:BirdResults}
\end{figure*}

The algorithm speeded-up turbo pixel (SUTP) computes super-pixels from a division of the image into rectangular regions (segments) of the same size, as shown in Fig.~\ref{Fig:Superpixels}. This approach does not take advantage of large uniform regions, for which the existence of many super-pixels is irrelevant. We propose a combined quadtree approach~\cite{quadtree} to generate the initial grid, followed by the SUTP algorithm. We call this method quadtree speeded-up turbo pixels (QSUTP). 
Figs.~ \ref{fig:SantaCatalinaResults}(b)--(d) and \ref{fig:BirdResults}(b)--(d) are the results obtained with the proposed QSUTP for original Figs.~\ref{fig:SantaCatalinaResults}(a) and \ref{fig:BirdResults}(a), respectively. The results obtained with the original SUTP are depicted in Figs.~\ref{fig:SantaCatalinaResults}(e)--(g) and \ref{fig:BirdResults}(e)--(g). We considered $\lambda_1=1$, $\lambda_2=0.09$ and radius$=5$ for both large images, as well as the adaptive thresholding approach, CIELAB colour model and the fast greedy community detection algorithm. For Figs.~\ref{fig:SantaCatalinaResults} and~\ref{fig:BirdResults}, SUTP generates accurate partitions for both original and quadtree approaches. However, QSUTP is more suitable than SUTP for large images, since it generates a small number of super-pixels.

Table \ref{Tab:Quadtree} shows the processing time for both experiments for each stage of the segmentation method, i.e. super-pixel extraction, graph generation and community detection. The super-pixel extraction uses most of the time required for image segmentation. QSTUP reduces the processing time significantly when images exhibit large uniform regions (see Fig.~\ref{fig:BirdResults}).

\begin{table*}[!htb]
\centering
\begin{tabular}{|c|c c c c|c c c c|}
\hline
& \multicolumn{4}{c|}{Original SUTP} & \multicolumn{4}{c|}{Quadtree SUTP}\\
%\cline{2-9}
 & SP & GG & FG & Total & SP & GG & FG & Total \\
\hline
Fig. \ref{fig:SantaCatalinaResults} & 1.21 & 1.03 & 0.25 & 2.49 & 1.69 & 0.59 & 0.14 & 2.42\\
Fig. \ref{fig:BirdResults}          & 2.01 & 3.27 & 1.95 & 7.23 & 1.45 & 0.99 & 0.33 & 2.77\\
\hline
\end{tabular}
\caption{Processing times (in seconds) for the original STUP and the proposed QSTUP, for each stage of the segmentation method, i.e. super-pixel computation (SP), graph generation (GG) and  Fast Greedy community detection (FG).}
\label{Tab:Quadtree}
\end{table*}

\subsection{Qualitative evaluation of the segmentation and the best parameter values}

To analyse how the radius and threshold parameters influence the accuracy of the image segmentation we considered the Berkeley dataset~\cite{ImgBerkeley} in our experiments. This dataset provides a different number of manual segmentations for each of the 300 images of size $481\times321$. Each image was segmented using radius values ranging from 1 to 5. The adaptive thresholding values ranged from 0.5 to 40.0, with a 0.5 increment. We adopted  $10\times 10$ initial size super-pixels, $\lambda_1=1$, $\lambda_2=0.1$ and the CIELAB colour model. Since we do not intend to penalize over-segmentation, we set the penalty parameter $\alpha=0$. As a result, we generated 400 segmentations for each image, reaching a total of 120,000 segmented images. To better compile the results, this massive amount of data was stored in a \textit{Postgresql}\footnote{http://www.postgresql.org/} database. These segmented images were compared with the reference images obtained by the method described in Section~\ref{subs:MetEvaluation}.

\begin{figure}[!t]
\begin{center}
\subfigure[]{\includegraphics[width=0.45\columnwidth]{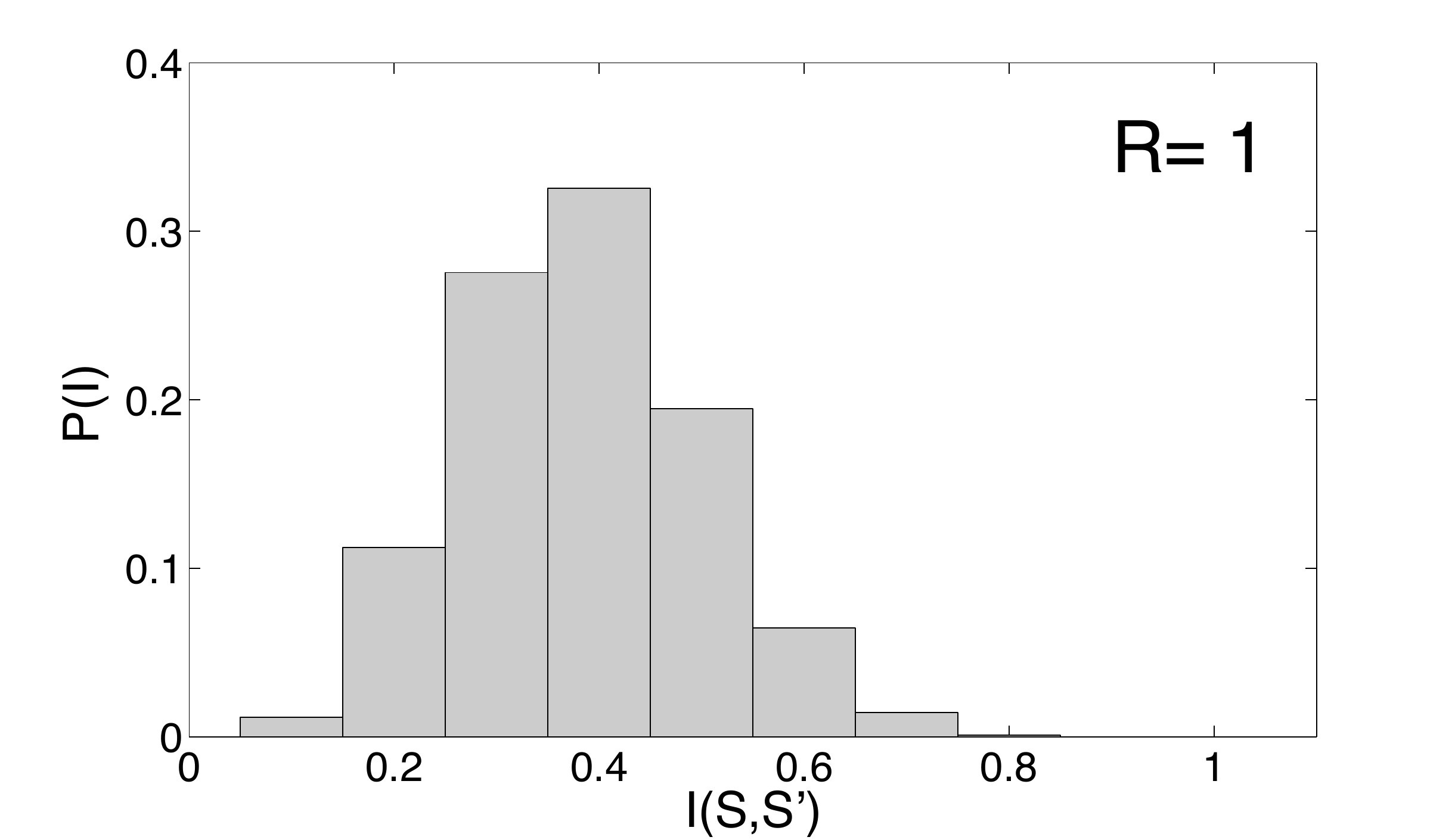}}
\subfigure[]{\includegraphics[width=0.45\columnwidth]{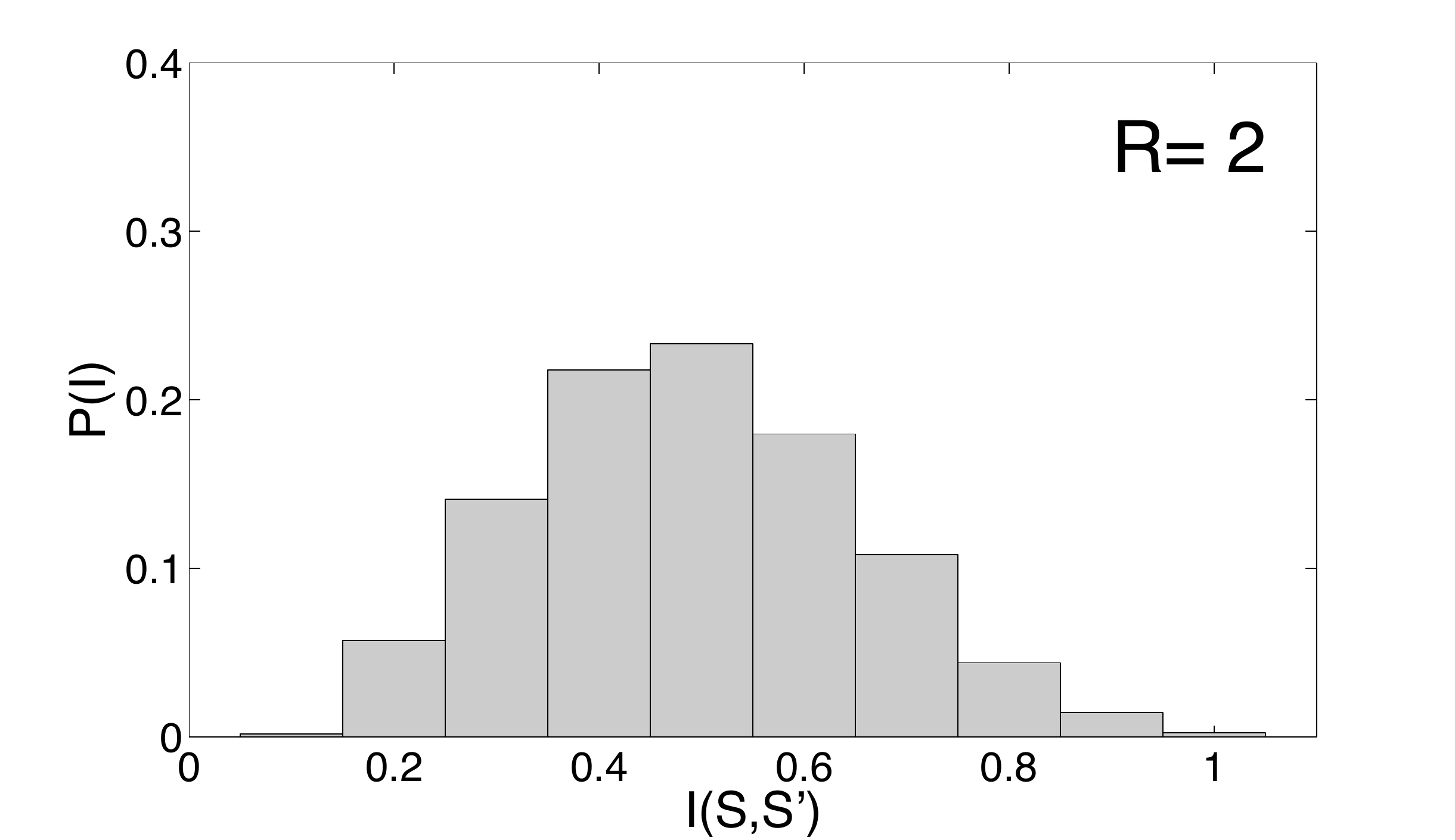}}
\subfigure[]{\includegraphics[width=0.45\columnwidth]{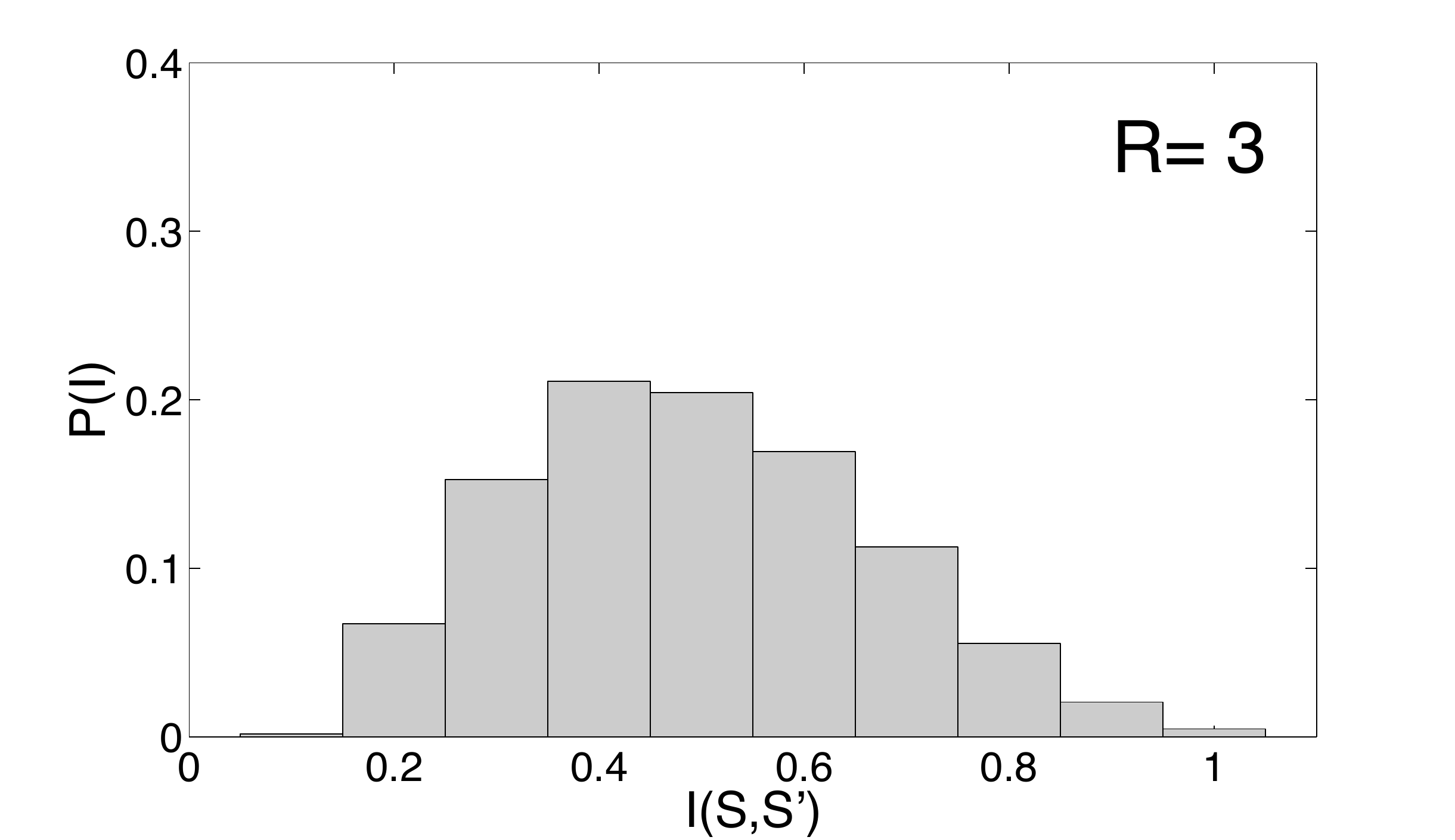}}
\subfigure[]{\includegraphics[width=0.45\columnwidth]{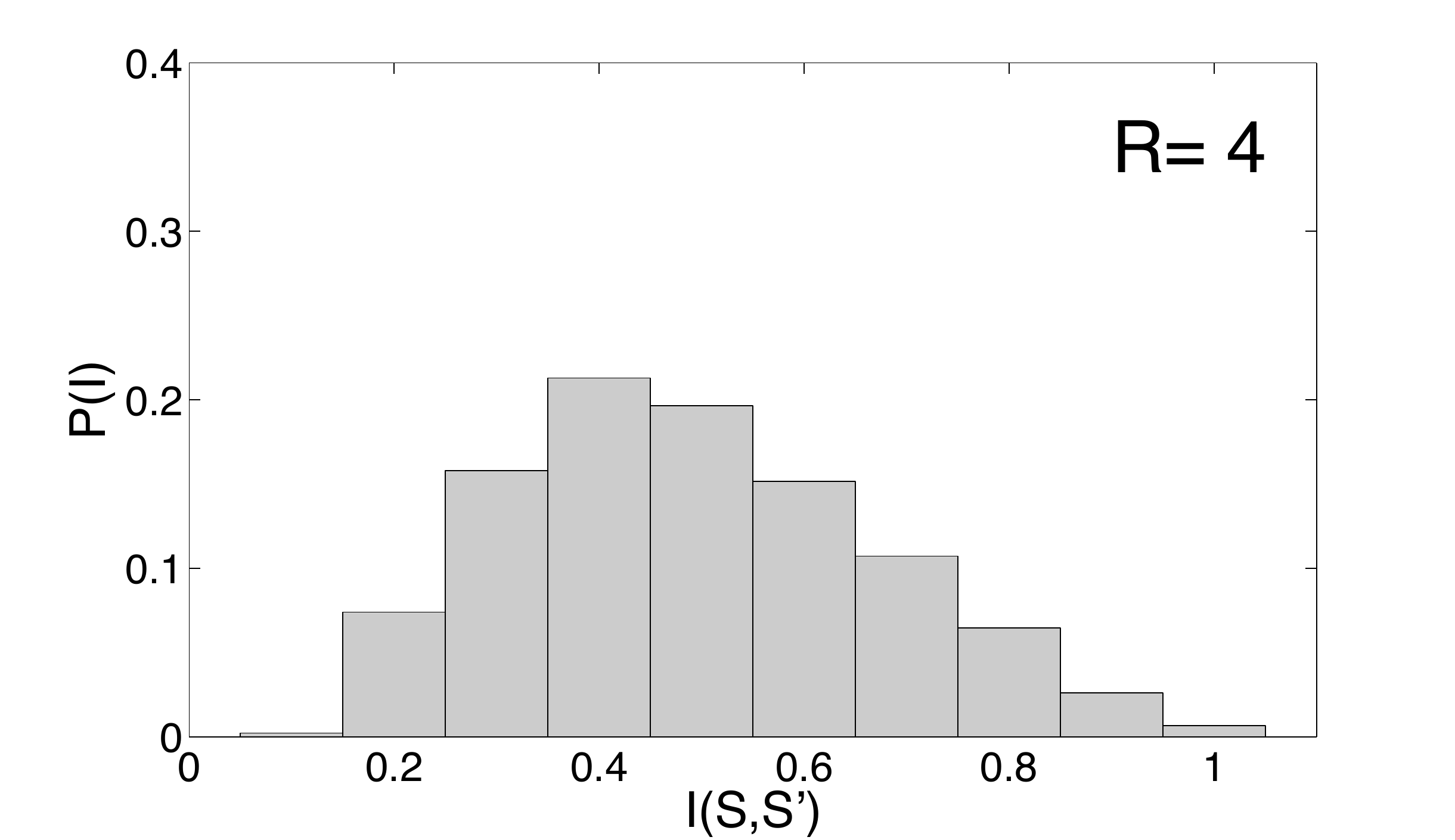}}
%\subfigure[]{\includegraphics[width=0.30\columnwidth]{r5.eps}}
\caption{Histograms of the segmentation quality $I(S,S')$ for (a) $R=1$, (b) $R=2$, (c) $R=3$ and (d) $R=4$.}
\label{Fig:HistogramsRadius}
\end{center}
\end{figure}

Fig.~\ref{Fig:HistogramsRadius} shows the histograms of the segmentation quality measure $I(S,S')$ - Eq.~\ref{eqn:IntersectionMetric} - for radius ranging from $R=1$ to $R=4$. Note that $0 \leq I(S,S') \leq 1$, with $I(S,S')=1$ representing the highest accuracy. The best quality was achieved for $R=4$. The mean value and standard deviation of the segmentation quality for $R=4$ are $\langle I\rangle = 0.74$ and $\sigma_I = 0.13$, respectively. For  $R=5$, $\langle I\rangle = 0.73$ and $\sigma_I = 0.14$. Given the similarity between $R=4$ or $R=5$, we inferred that either can be used for an accurate segmentation. Although there was no quality loss for $R>5$,  the computational costs increased.

\begin{figure}[!t]
\begin{center}
\subfigure[]{\includegraphics[width=0.45\columnwidth]{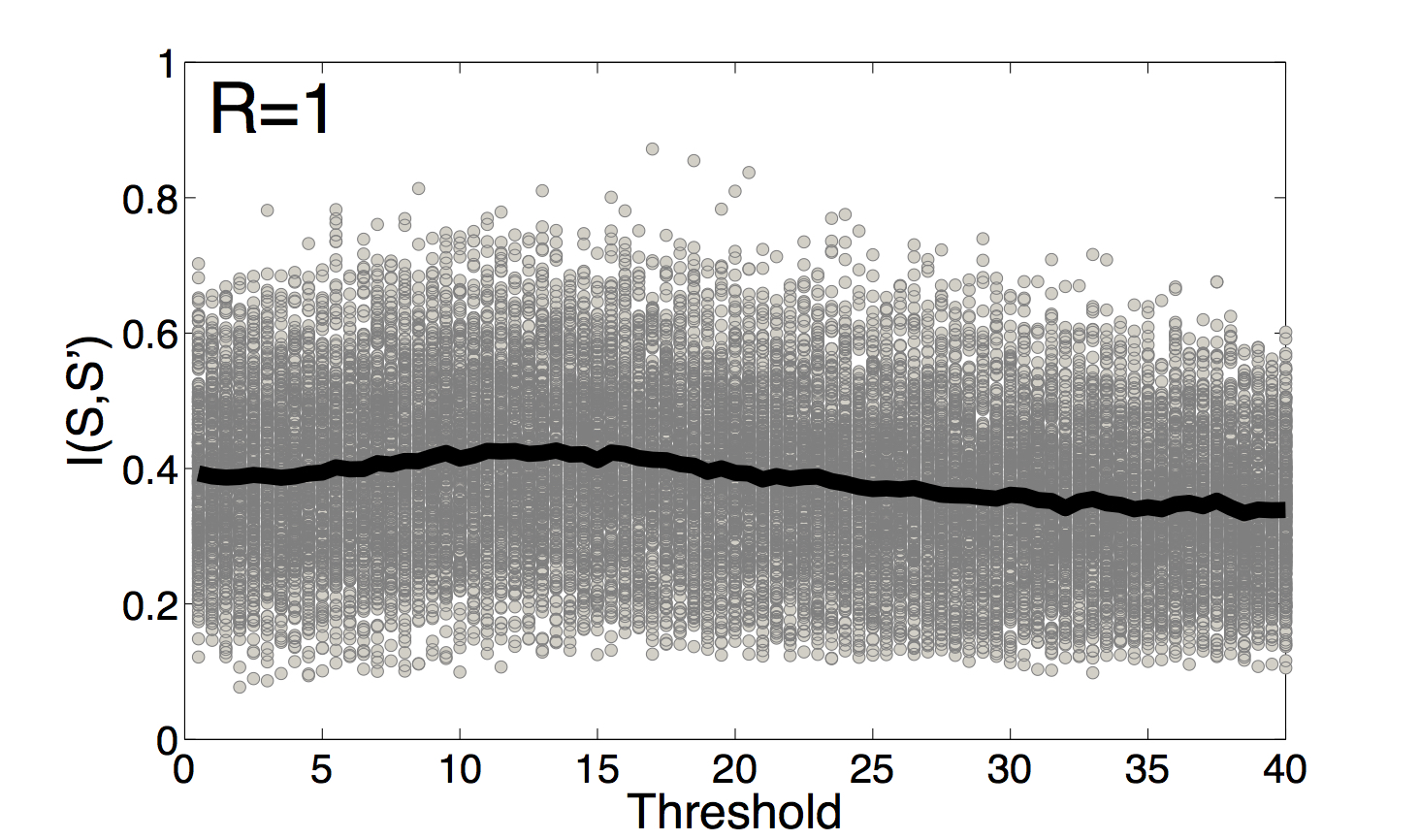}}
\subfigure[]{\includegraphics[width=0.45\columnwidth]{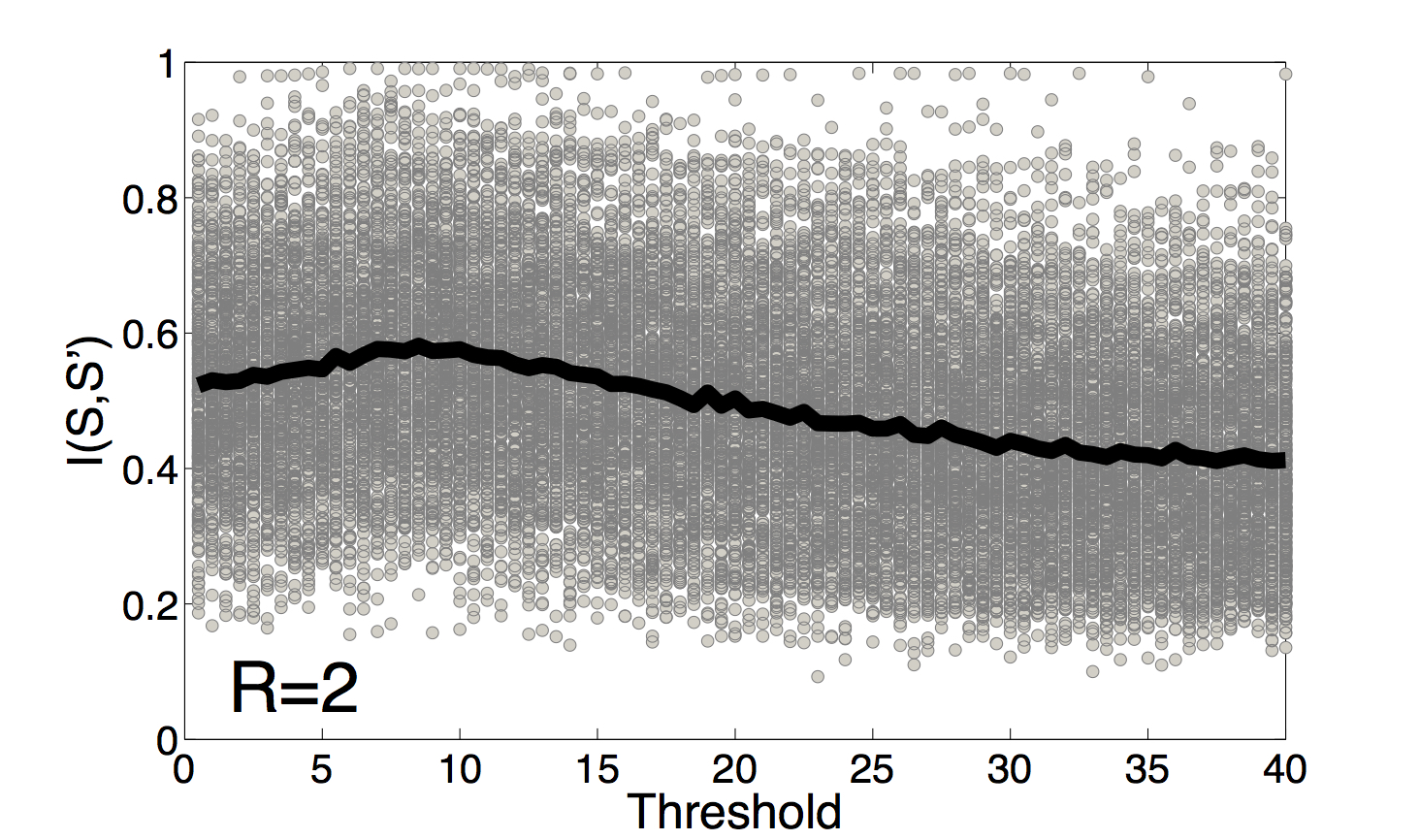}}
\subfigure[]{\includegraphics[width=0.45\columnwidth]{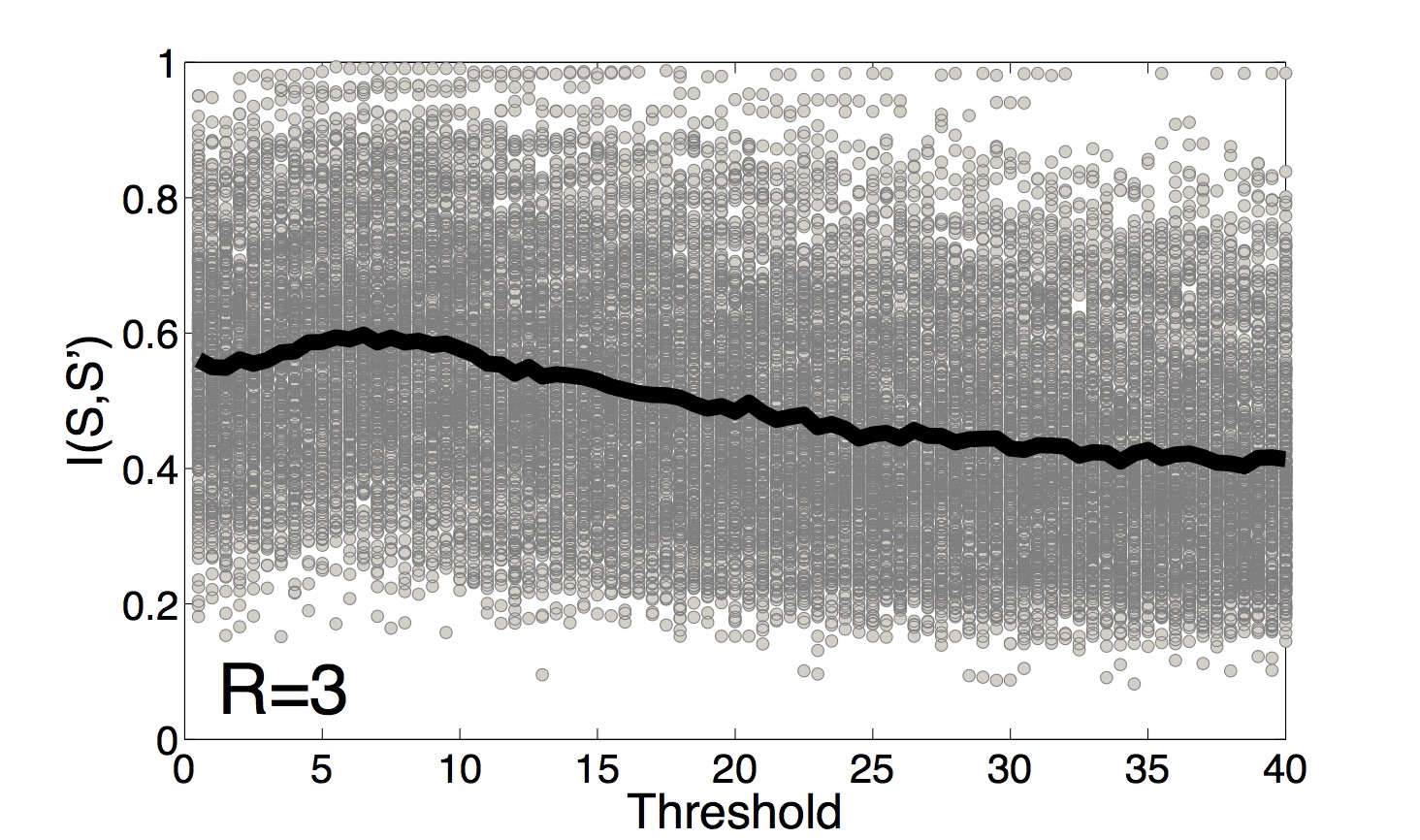}}
\subfigure[]{\includegraphics[width=0.45\columnwidth]{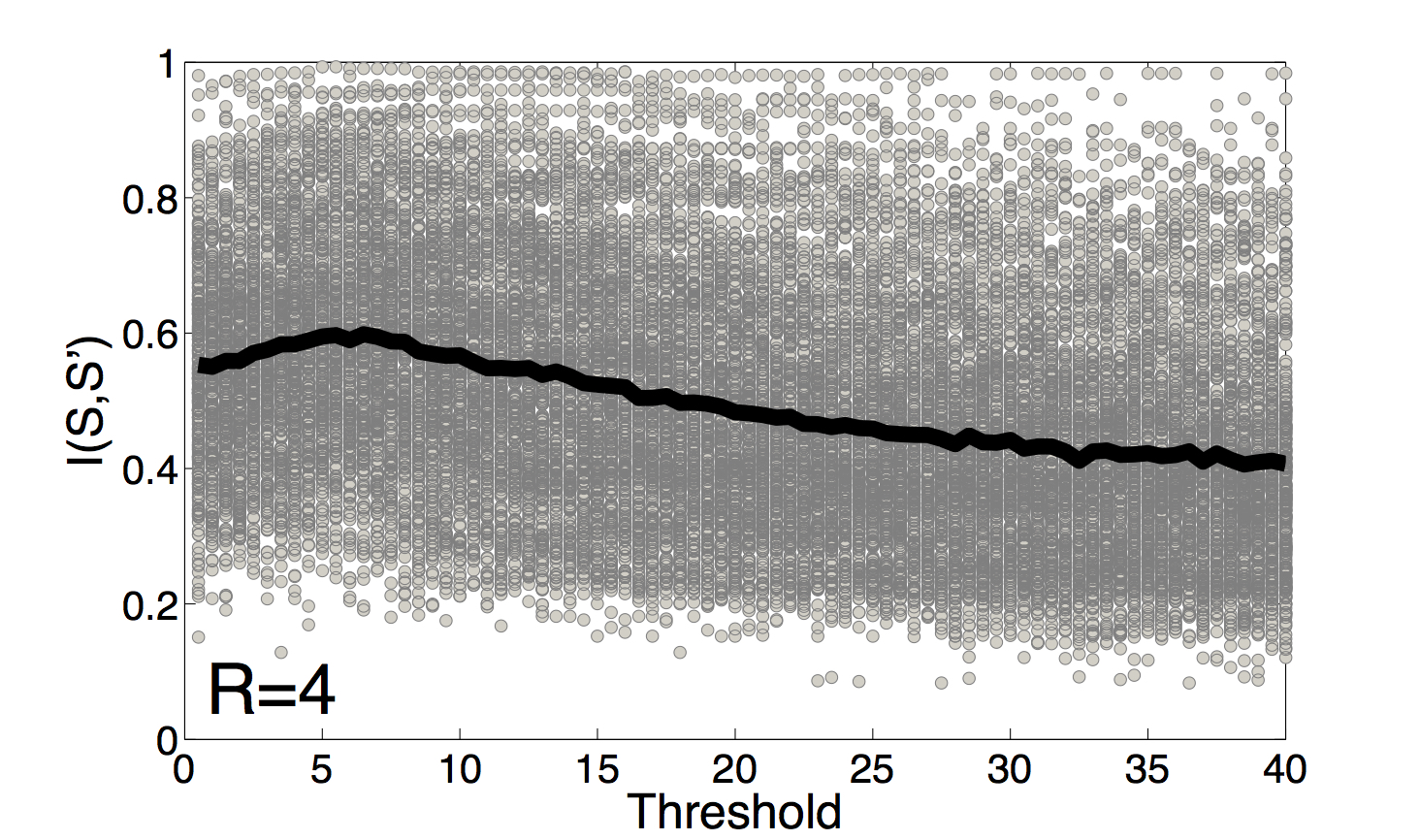}}
\caption{Relationship between $I(S,S')$ and the threshold for (a) $R=1$, (b) $R=2$, (c) $R=3$ and (d) $R=4$.}
\label{Fig:HistogramsThreshold}
\end{center}
\end{figure}

\begin{figure}[!tb]
\centering
\includegraphics[width=.8\linewidth]{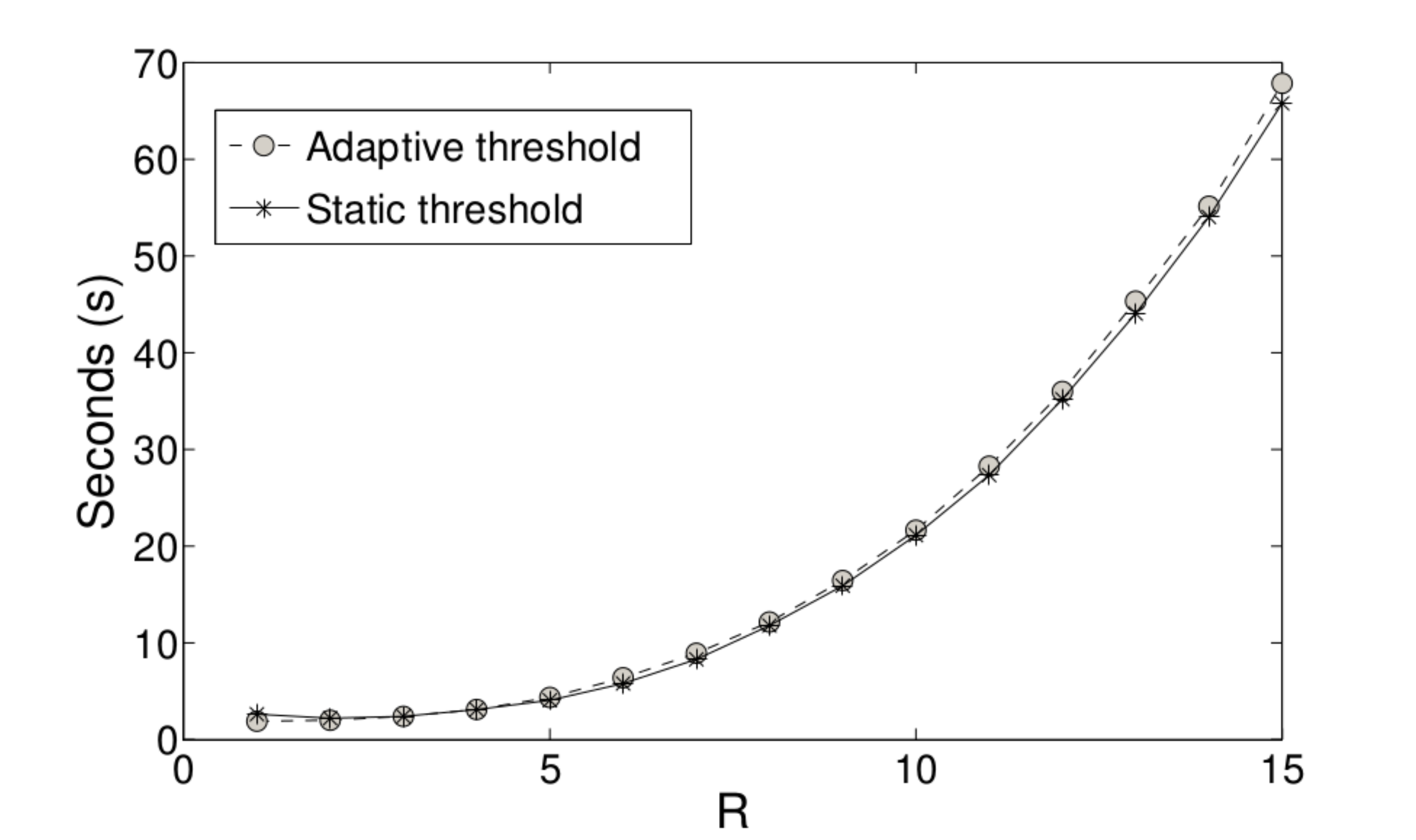}
  \caption{Processing times (in seconds) for increasing values of radius $R$: static ($t=0.75$) and adaptive thresholds.}
 \label{Fig:RadiusTime}
\end{figure}

Fig.~\ref{Fig:HistogramsThreshold} shows the relationship between threshold ($t$) and the quality of the image segmentation for different values of $R$. We considered an adaptive thresholding scheme, with $t$ ranging from 0.5 to 40, with a 0.5 increment at each iteration. The Pearson correlation coefficient between the quality of the segmentation $I(S,S')$ and the threshold is $\rho \approx -0.3$ for all cases. Unlike the radius, the threshold does not have a significant influence on the segmentation results. The average quality of the segmentation suggests that the most accurate segmentations can be obtained for $t < 15$. The plot in Fig.~\ref{Fig:RadiusTime} shows that the processing time for segmentation equally increases with the radius for both static and adaptive thresholding. The use of the adaptive thresholding approach does not imply a significant increase in processing times, even when some super-pixels are evaluated more than once. Moreover, the use of the adaptive threshold provided more accurate results.

\begin{figure*}[!htb]
\centering
  \subfigure[Original]{\includegraphics[width=.30\linewidth]{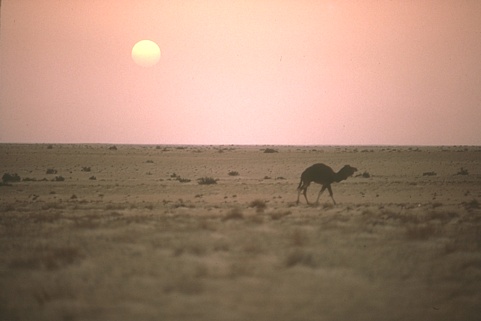}}
  \subfigure[Original]{\includegraphics[width=.30\linewidth]{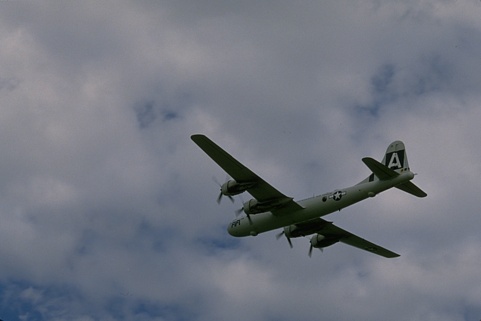}}
  \subfigure[Original]{\includegraphics[width=.30\linewidth]{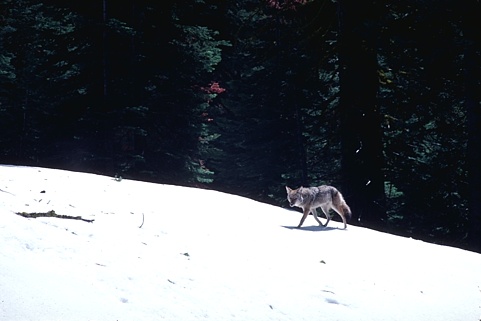}} \\
  \subfigure[Ground Truth]{\includegraphics[width=.30\linewidth]{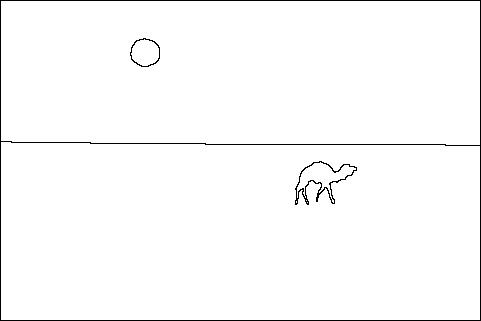}}
  \subfigure[Ground Truth]{\includegraphics[width=.30\linewidth]{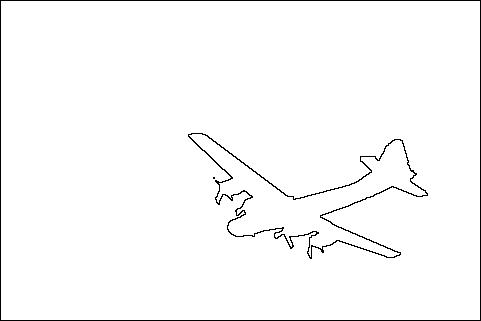}}
  \subfigure[Ground Truth]{\includegraphics[width=.30\linewidth]{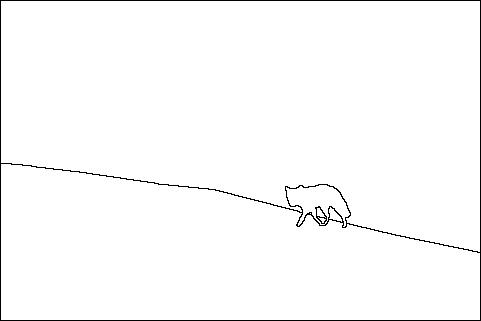}} \\
  \subfigure[$I(S,S')=0.99$]{\includegraphics[width=.30\linewidth]{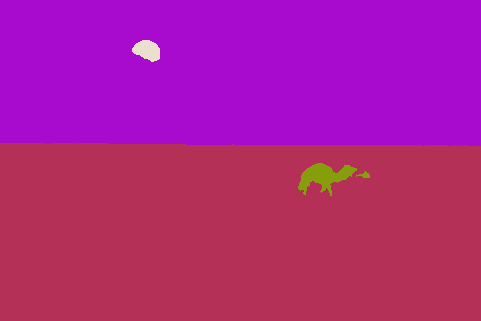}}
  \subfigure[$I(S,S')=0.98$]{\includegraphics[width=.30\linewidth]{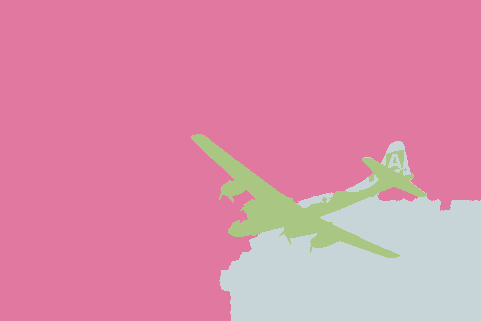}}
  \subfigure[$I(S,S')=0.99$]{\includegraphics[width=.30\linewidth]{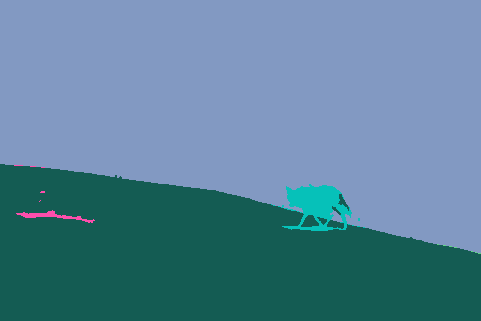}}\\
  \caption{Segmentation produced by the proposed methodology  based on super-pixels and community detection in graphs.}
 \label{Fig:Results}
\end{figure*}

Figs.~ \ref{Fig:Results} and ~\ref{Fig:Comparison} show the results of our segmentation approach for some natural scene images. Segmentations are very accurate in comparison with the manual segmentation of the Berkeley dataset. Recall that segmentation quality is computed by taking the reference image obtained as described in Section~\ref{subs:MetEvaluation}, which can be understood as the most likely human segmentation.

\subsection{Comparison with related image segmentation methods}

We compared our approach with two segmentation methods: a) Felzenszwalb and Huttenlocher's \cite{Felzenswalb} and Arbelaez's \cite{arbelaez2011contour}. The former is a graph-based approach which is suitable for large images due to its low computational costs (see Section~\ref{sec:RelatedWork}). The latter is a contour-based image segmentation method with multi-scale capabilities which delivers highly accurate results. Quantitative analysis presented in this section is the result of segmentation accuracy (according to the metric presented in \cite{Cuadros012})  for all the 300 images on the Berkeley image database. The segmentation quality mean and the standard deviation for our method were $\langle I\rangle = 0.74$ and $\sigma_I = 0.13$ for $R=4$, respectively. We summarize in Table \ref{tb_comparison} the results produced by  Felzenszwalb and Huttenlocher, Arbelaez and our method. 
\begin{table}[h]
\centering
\begin{tabular}{|c|c|c|c|}
	\hline 
\textbf{Method}	& \textbf{Mean} & \textbf{Standard Deviation } \\ 
	\hline 
Felzenszwalb	& 0.48  & 0.52   \\ 
	\hline 
Arbelaez	& 0.67 & 0.16 \\ 
	\hline 
Community Detection	& 0.74 & 0.13 \\ 
	\hline 
\end{tabular} 
\caption{Segmentation quality Mean and standard deviation computed for the Berkeley database.}
\label{tb_comparison}
\end{table}

\subsubsection{Felzenszwalb and Huttenlocher} 

\begin{figure}[!htb]
\centering
  \subfigure[Original]{\includegraphics[width=.24\linewidth]{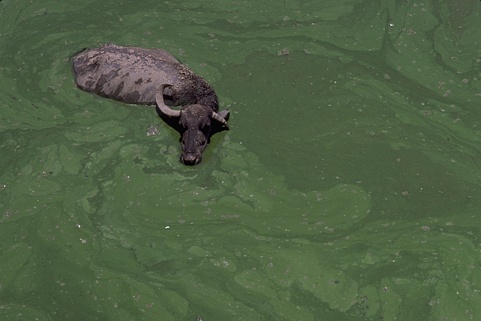}}
  \subfigure[Ground Truth]{\includegraphics[width=.24\linewidth]{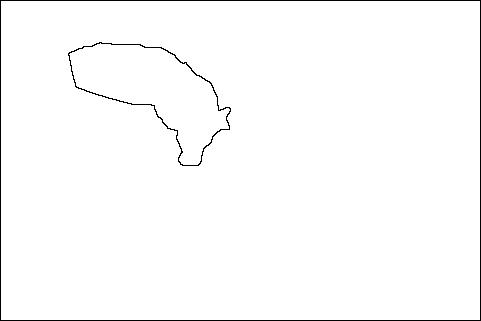}}
  \subfigure[$I(S,S') =0.98$]{\includegraphics[width=.24\linewidth]{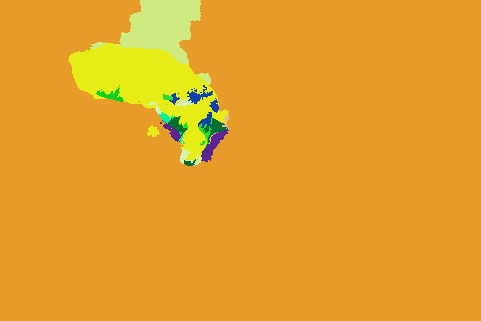}}
  \subfigure[$I(S,S') =0.94$]{\includegraphics[width=.24\linewidth]{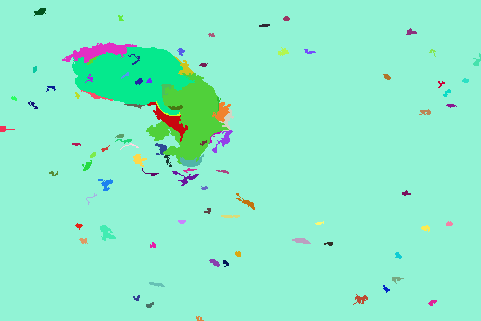}} \\

  \subfigure[Original]{\includegraphics[width=.24\linewidth]{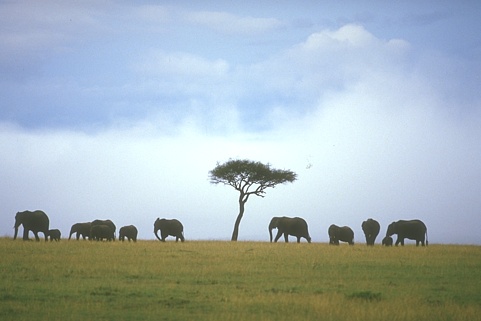}}
  \subfigure[Ground Truth]{\includegraphics[width=.24\linewidth]{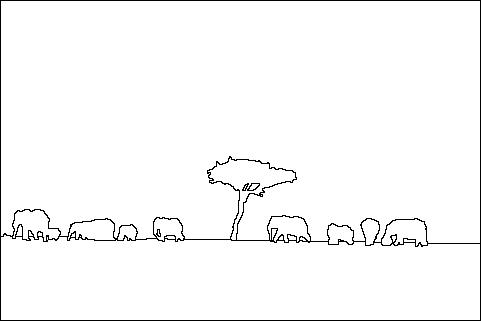}}
  \subfigure[$I(S,S') =0.97$]{\includegraphics[width=.24\linewidth]{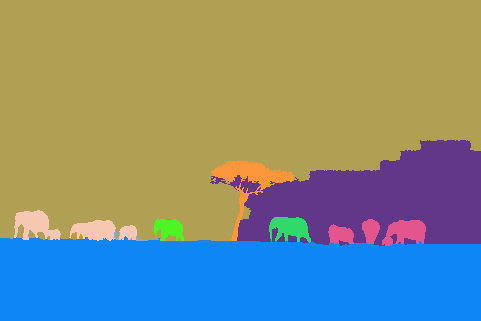}}
  \subfigure[$I(S,S') =0.94$]{\includegraphics[width=.24\linewidth]{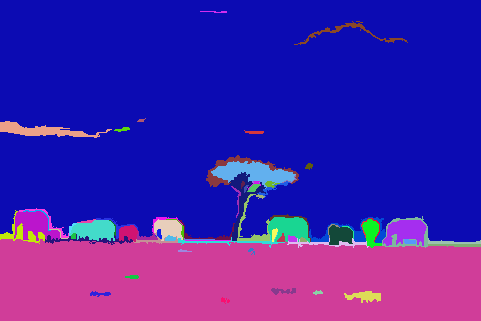}}\\

  \subfigure[Original]{\includegraphics[width=.24\linewidth]{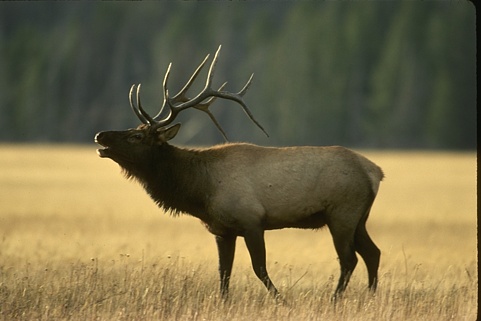}}
  \subfigure[Ground Truth]{\includegraphics[width=.24\linewidth]{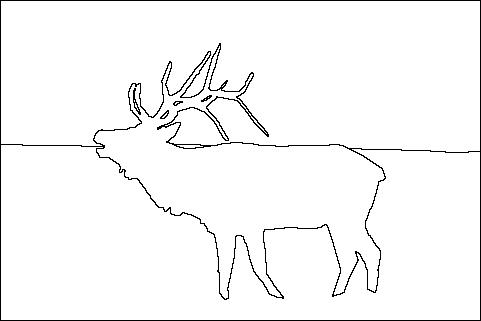}}
  \subfigure[$I(S,S') =0.96$]{\includegraphics[width=.24\linewidth]{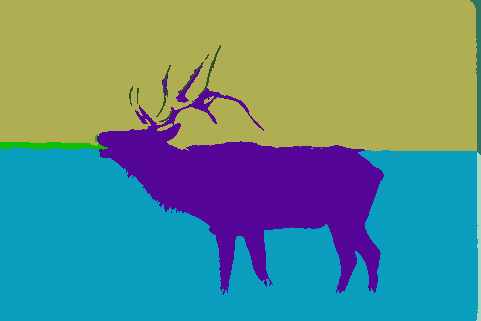}}
  \subfigure[$I(S,S') =0.65$]{\includegraphics[width=.24\linewidth]{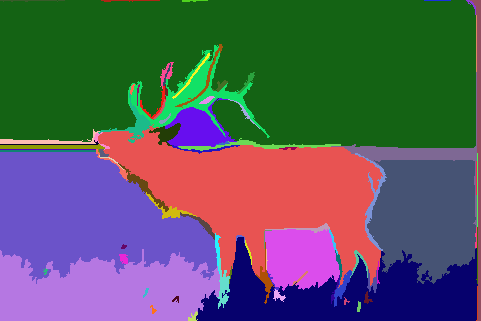}}\\

  \subfigure[Original]{\includegraphics[width=.24\linewidth]{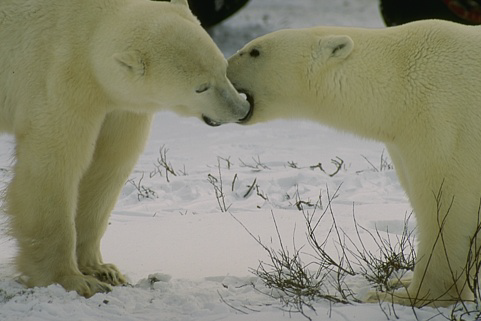}}
  \subfigure[Ground Truth]{\includegraphics[width=.24\linewidth]{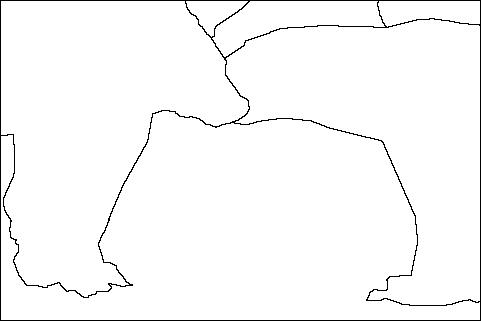}}
  \subfigure[$I(S,S') =0.95$]{\includegraphics[width=.24\linewidth]{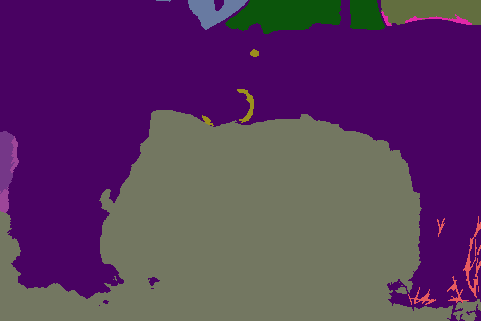}}
  \subfigure[$I(S,S') =0.77$]{\includegraphics[width=.24\linewidth]{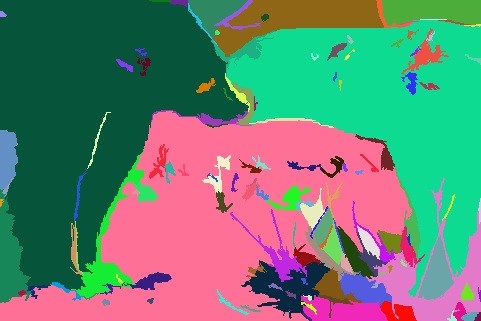}}\\
  
  \caption{Original and ground truth images are in the first and second columns, respectively. Segmentations produced by our method and Felzenszwalb and Huttenlocher's~\cite{Felzenswalb} are in the third and fourth columns, respectively. $I(S,S')$ indicates the quality of the segmentation according to our evaluation technique given in Section~\ref{subs:MetEvaluation}.}
 \label{Fig:Comparison}
\end{figure}

 We used the parameters as described in the original paper. The mean and standard deviation of the segmentation quality for the Berkeley database were $\langle I\rangle = 0.48$ and $\sigma_I = 0.52$, respectively. Note that this mean value is close to the random segmentation. Fig.~ \ref{Fig:Comparison} shows some segmentations obtained with our approach and the algorithm by Felzenszwalb and Huttenlocher. In all cases our method provided more accurate results.

\subsubsection{Arbelaez's Contour-based Image Segmentation} 

\begin{figure}[!htb]
	\centering
	\subfigure[]{\includegraphics[width=.24\linewidth]{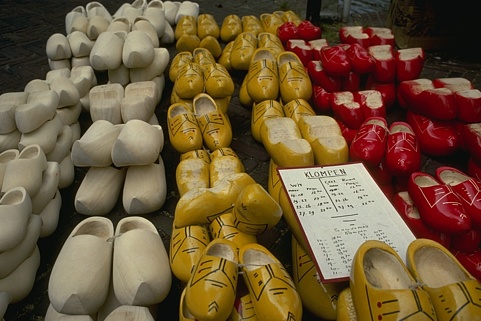}}
	\subfigure[Ground Truth]{\includegraphics[width=.24\linewidth]{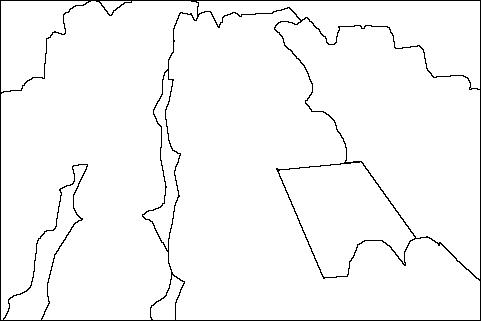}}
	\subfigure[$I(S,S') =0.79$]{\includegraphics[width=.24\linewidth]{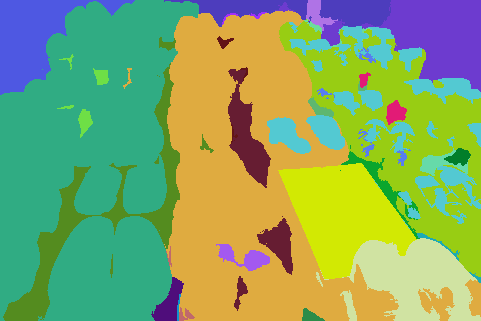}}
	\subfigure[$I(S,S') =0.66$]{\includegraphics[width=.24\linewidth]{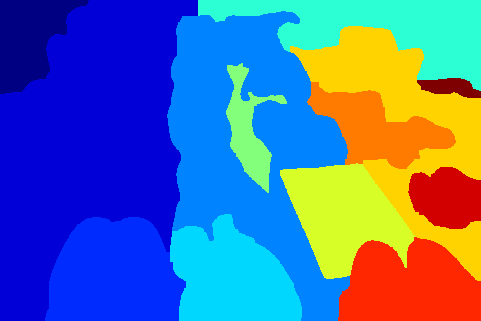}} \\
	
	\subfigure[]{\includegraphics[width=.24\linewidth]{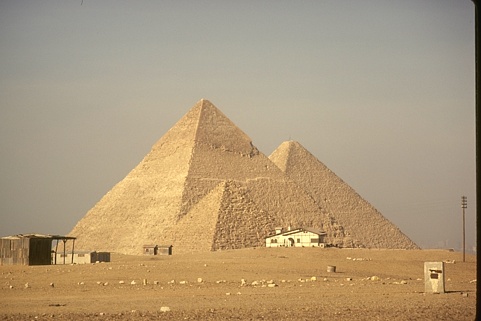}}
	\subfigure[Ground Truth]{\includegraphics[width=.24\linewidth]{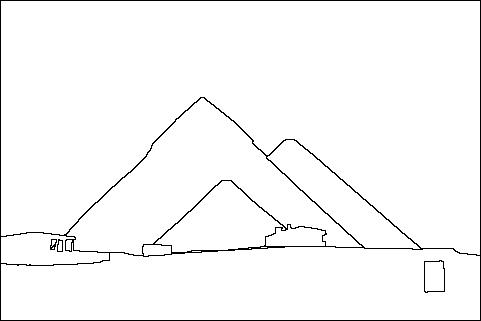}}
	\subfigure[$I(S,S') =0.78$]{\includegraphics[width=.24\linewidth]{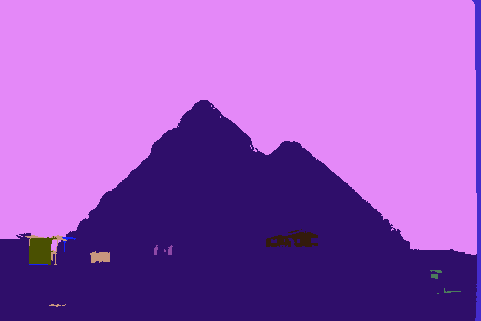}}
	\subfigure[$I(S,S') =0.77$]{\includegraphics[width=.24\linewidth]{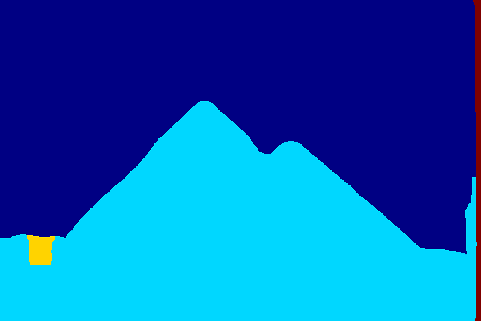}} \\
	
	\subfigure[]{\includegraphics[width=.24\linewidth]{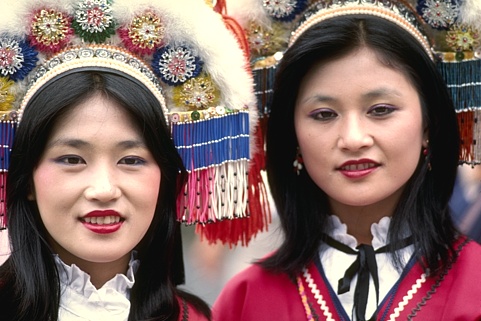}}
	\subfigure[Ground Truth]{\includegraphics[width=.24\linewidth]{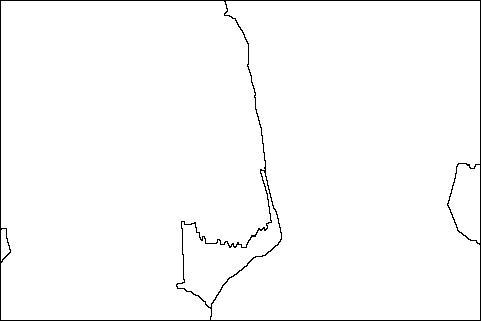}}
	\subfigure[$I(S,S') =0.53$]{\includegraphics[width=.24\linewidth]{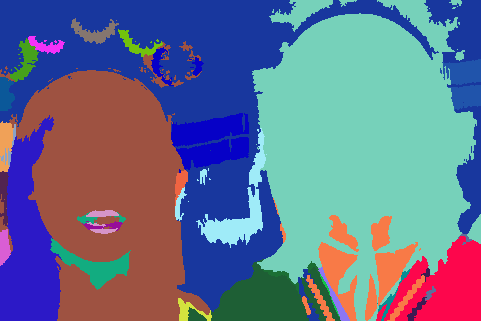}}
	\subfigure[$I(S,S') =0.37$]{\includegraphics[width=.24\linewidth]{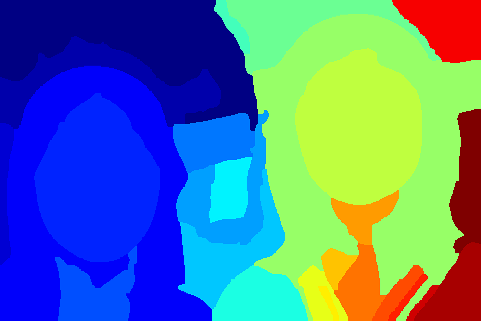}} \\
	
	\subfigure[]{\includegraphics[width=.24\linewidth]{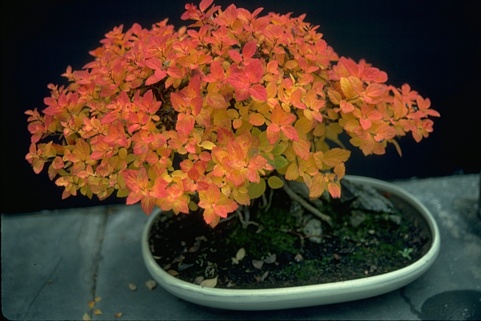}}
	\subfigure[Ground Truth]{\includegraphics[width=.24\linewidth]{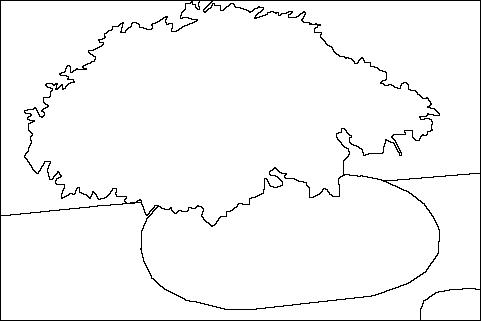}}
	\subfigure[$I(S,S') =0.88$]{\includegraphics[width=.24\linewidth]{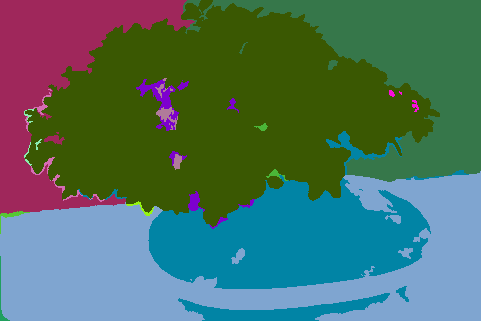}}
	\subfigure[$I(S,S') =0.91$]{\includegraphics[width=.24\linewidth]{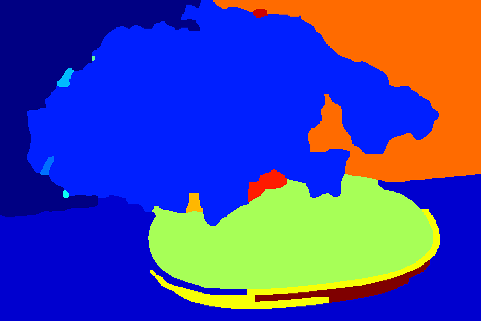}} \\
	
	\caption{Original and ground truth images are in the first and second columns, respectively. Segmentations produced by our method and  Arbelaez's~\cite{arbelaez2011contour} are in the third and fourth columns, respectively.  $I(S,S')$ indicates the quality of the segmentation according to our evaluation technique given in Section~\ref{subs:MetEvaluation}.}
	\label{Fig_arbelaez_comp}
\end{figure}

We performed a similar experiment with Arbelaez's method, running the available source code implemented by the authors. The mean and standard deviation of the segmentation quality for the Berkeley database were $\langle I\rangle = 0.67$ and $\sigma_I = 0.16$, respectively. Figure \ref{Fig_arbelaez_comp} shows the segmentation results of both proposed approach and Abelaez's for 4 samples selected from the database. Notice that the accuracy of our method is slightly better, but the processing average time per image is much superior. Our method takes 4 seconds whereas Arbelaez's takes 90 seconds in average.

\section{Conclusions} \label{sec:Conclusion}

The current article has reported on a framework for the segmentation of large images based on super-pixels and community detection in graphs. We conducted a statistical analysis with 120,000 segmented images obtained from the Berkeley dataset. A new quantitative evaluation technique for image segmentation has also been introduced. A technique that automatically selects reference images, given a set of manually segmented images, has been implemented. We modified the super-pixel algorithm by Cigla and Alatan~\cite{Cigla} in order to replace the original regular grid approach for a quadtree-based grid. Quadtree contributes to reduce the number of super-pixels when images have large and uniform regions. The fast greedy community detection algorithm for graph partition was employed. Results show that our method is faster and more accurate than the algorithm by Felzenszwalb and Huttenlocher~\cite{Felzenswalb}, which is also based on graph concepts. It is also more accurate than Arbelaez's, although by a narrower margin. However, our approach is much faster. This is due to super-pixels, which speed up the process and allow the segmentation of large images in a much faster way.     

\section{Acknowledgments}

The authors would like to acknowledge the support of Brazilian research agencies S{\~a}o Paulo Research Foundation (FAPESP) (Process number 2011/05802-2) and CNPQ. F.A.R. also would like to acknowledge NAP eScience - PRP -- USP. The authors also thank Angela C. P. Giampedro, who provided a careful review of the text.

\bibliographystyle{unsrt}      % basic style, author-year citations
\bibliography{references}   % name your BibTeX data base

\end{document}